\documentclass[acmtog]{acmart}
\setcopyright{rightsretained}
\acmJournal{TOG}
\acmYear{2023} \acmVolume{1} \acmNumber{1} \acmArticle{1} \acmMonth{1} \acmPrice{}\acmDOI{10.1145/3592460}

\acmSubmissionID{896}

\usepackage{booktabs} %

\usepackage{graphicx}
\usepackage{bm}
\usepackage{xcolor}
\usepackage{microtype} 
\usepackage{subcaption}

\usepackage[ruled]{algorithm2e} %

\SetAlFnt{\small}
\SetAlCapFnt{\small}
\SetAlCapNameFnt{\small}
\SetAlCapHSkip{0pt}

\acmJournal{TOG}

\usepackage[capitalize]{cleveref}
\crefname{section}{Sec.}{Secs.}
\Crefname{section}{Section}{Sections}
\Crefname{table}{Table}{Tables}
\crefname{table}{Tab.}{Tabs.}

\usepackage{ifthen}
\newboolean{revisionMode}
\setboolean{revisionMode}{false} %
\newcommand{\revision}[1]{\ifthenelse{\boolean{revisionMode}}{\textcolor{blue}{#1}}{#1}}

\begin{document}
\title{Real-Time Radiance Fields for Single-Image Portrait View Synthesis}

\author{Alex Trevithick}
\affiliation{%
  \institution{University of California San Diego}
  \city{La Jolla}
  \country{USA}}
    \authornote{This project was initiated and substantially carried out during an internship at NVIDIA.}

\author{Matthew Chan}
\author{Michael Stengel}
\affiliation{%
  \institution{NVIDIA}
  \city{Santa Clara}
  \country{USA}}

\author{Eric R. Chan}
\authornotemark[1]
\affiliation{%
  \institution{Stanford University}
  \city{Stanford}
  \country{USA}}

\author{Chao Liu}
\author{Zhiding Yu}
\author{Sameh Khamis}
\affiliation{%
  \institution{NVIDIA}
  \city{Santa Clara}
  \country{USA}}

\author{Manmohan Chandraker}
\author{Ravi Ramamoorthi}
\affiliation{%
  \institution{University of California San Diego}
  \city{La Jolla}
  \country{USA}}

\author{Koki Nagano}
\affiliation{%
  \institution{NVIDIA}
  \city{Santa Clara}
  \country{USA}}

\begin{abstract}
We present a one-shot method to infer and render a photorealistic 3D representation from a single unposed image (e.g., face portrait) in real-time. Given a single RGB input, our image encoder directly predicts a canonical triplane representation of a neural radiance field for 3D-aware novel view synthesis via volume rendering. Our method is fast (24 fps) on consumer hardware, and produces higher quality results than strong GAN-inversion baselines that require test-time optimization. To train our triplane encoder pipeline, we use only synthetic data, showing how to distill the knowledge from a pretrained 3D GAN into a feedforward encoder. Technical contributions include a Vision Transformer-based triplane encoder, a camera data augmentation strategy, and a well-designed loss function for synthetic data training. We benchmark against the state-of-the-art methods, demonstrating significant improvements in robustness and image quality in challenging real-world settings. We showcase our results on portraits of faces (FFHQ) and cats (AFHQ), but our algorithm can also be applied in the future to other categories with a 3D-aware image generator.
\end{abstract}




\begin{teaserfigure}
\begin{center}
    \centering
    \captionsetup{type=figure}
    \includegraphics[width=\textwidth]{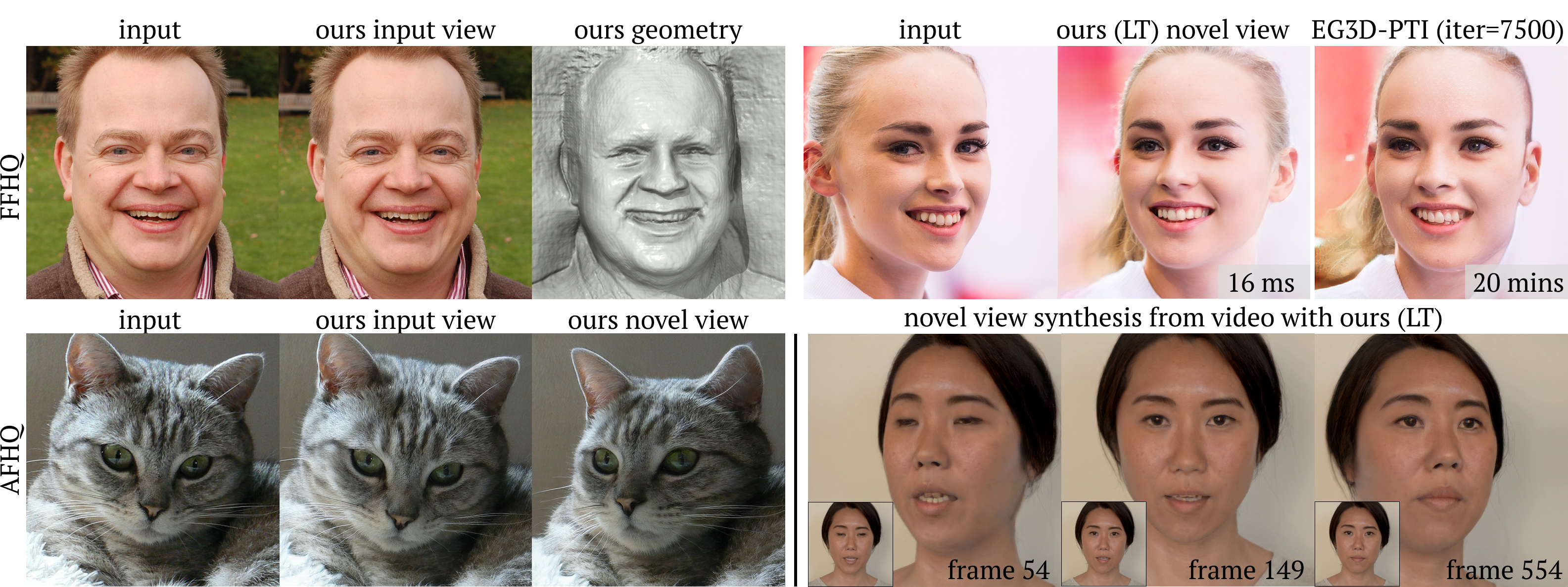}
    \captionof{figure}{
  Given a single RGB input image, our method generates 3D-aware images and geometry of an object (e.g., faces [top row] and cats [bottom row, left]) in real-time, while the state-of-the-art 3D GAN inversion~\cite{eg3d2022} does not generate a satisfactory result after 20 mins of fine-tuning~\cite{roich2021pivotal} (top right). Our method can also be applied to a video frame-by-frame for video-based novel view synthesis (bottom row, right). Ours (LT) refers to a lightweight faster version of our 
  model that has almost the same quality as the full model.
  Credits to Erik (HASH) Hersman and 2017 Canada Summer Games.}
    \label{fig:teaser}
\end{center}
\end{teaserfigure}

\maketitle

\begin{figure*}
    \centering
    \includegraphics[width=\linewidth]{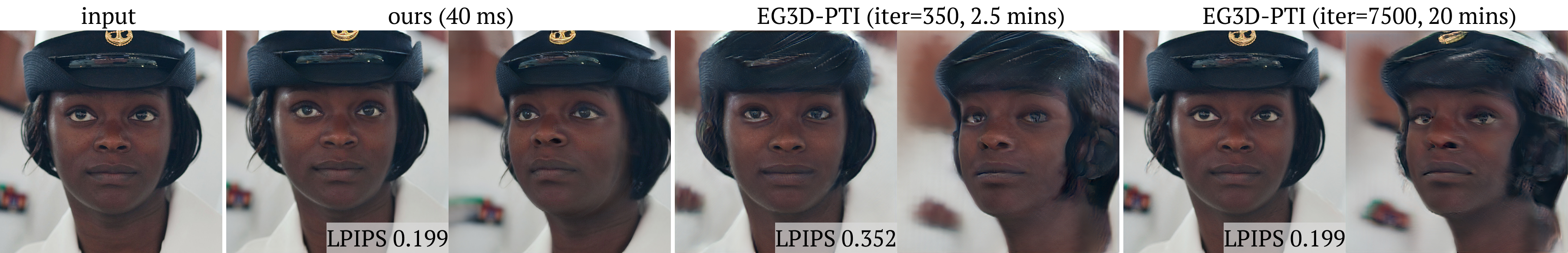}
    \caption{Comparison to the state-of-the-art 3D GAN\cite{eg3d2022} with test-time fine tuning\cite{roich2021pivotal} (EG3D-PTI). Single-view 3D GAN inversion approaches trade off the 2D reconstruction quality and the 3D effects. When fine tuned longer (7500 iterations), EG3D-PTI can capture the same fine-scale details as ours ($LPIPS=0.199$), but the quality of another view starts to degrade. On the other hand, our method captures out-of-domain details (e.g., emblem) in one-shot while producing realistic rendering of another view, and operating in real-time. Credit to Obama White House.}
    \label{fig:pti_comparison}
\end{figure*} 

\section{Introduction}
\label{sec:intro}

Digitally reproducing the 3D appearance of an object from a single image is a long-standing goal for computer graphics and vision. Interactive synthesis of photorealistic novel views opens new possibilities for AR/VR, and for 3D telepresence and videoconferencing when applied to humans.  In this work, we propose a technique to infer a 3D representation for real-time view synthesis given a single portrait-style input image (e.g., of a human face, see Fig.~\ref{fig:teaser}).

Recently, 3D aware-image generation approaches (e.g.,~\cite{eg3d2022,deng2022gram,epigraf}) demonstrated unconditional generation of photorealistic 3D representations from a collection of single-view 2D images by combining NeRF-based representations~\cite{mildenhall2020nerf} and GANs~\cite{goodfellow2014generative}. Notably, EG3D~\cite{eg3d2022} proposed an efficient triplane 3D representation and demonstrated real-time 3D-aware image rendering with quality comparable to 2D GANs. Once trained, the 3D GAN generators can be frozen and used for single-image 3D reconstruction tasks via GAN inversion~\cite{Karras2020stylegan2} and test-time fine tuning~\cite{roich2021pivotal}. 
However, there are a few challenges in this 3D-GAN inversion-based methods. (1) Due to the multi-view nature of training a NeRF, it needs careful optimization objectives and additional 3D priors~\cite{yin2022spi,xie2022high} in the single view setting to avoid unsatisfactory results on novel views and corrupted geometry (see Fig. \ref{fig:comparison_1}). Fig.~\ref{fig:pti_comparison} shows the tradeoff in the SOTA single-view 3D GAN inversion pipeline. (2) The test-time optimization requires an accurate camera pose as input or to be jointly optimized~\cite{ko20233d3dganinversion}. (3) The above optimization for every single image is time-consuming, limiting the technique for real-time video applications.

In this paper, we %
present a one-shot approach to lift an input 2D portrait image to 3D in real-time (24fps on consumer hardware, see Tab.~\ref{tab:runtime}). Unlike previous work that reuses a pre-trained generator, we train an encoder end-to-end that directly predicts the triplane 3D features from a single input image. In contrast to prior works that use multiview real image acquisition setups, we do not need any real images at all, nor do we require time-consuming physically-based rendering of high-quality and expensive face assets.

Instead, we fully supervise the training of our triplane encoder for novel view synthesis using multiview-consistent synthetic data generated from a pre-trained 3D GAN. Together with our data augmentation strategies and Transformer-based encoder, we present a model which can handle challenging real-world input images including occlusion and three-quarter views. %
We showcase our results on human and cat face categories in this paper, but the methodology can apply to any category for which 3D-aware image generators 
are available. Our work may motivate applications such as temporally consistent 
view synthesis; Fig.~\ref{fig:teaser} (bottom right) shows our method applied to a 
video in a frame-by-frame fashion without any special handling.

In summary, contributions of our work include:
\begin{itemize}
    \item We propose a feed forward encoder model to directly infer a triplane 3D representation from an input image. No test-time optimization is needed.
    \item We present a new strategy for training a feed forward triplane encoder for 3D inversion using \textit{only} synthetic data generated from a pre-trained 3D-aware image generator.  %
    \item We demonstrate that our method can infer a photorealistic 3D representation in real-time given a single \textit{unposed} image. Together with our Transformer-based encoder and on-the-fly augmentation strategy, our method can robustly handle challenging input images of side views and occlusions. 
\end{itemize}

\begin{table}[t]
\setlength{\tabcolsep}{4.5pt}
    \centering
		\small
	  \caption{Time taken to lift the input image to 3D (Encoding) and render (Render) a 3D representation given an input image on a single RTX 3090 GPU. The end-to-end runtime with our model and our lightweight model (LT) is significantly faster than NeRF-based baselines. $^\dag$ROME employs 2D-based neural rendering with mesh-based neural textures, producing the output at 256x256 resolution; it also requires a segmentation mask and detected keypoints from off-the-shelf models which requires around 200ms. %
    }
    \begin{tabular}{@{\hskip 1mm}l c c c c c@{\hskip 1mm}}
		\toprule
		Time & H.NeRF & ROME & EG3D-PTI & Ours & Ours (LT)\\
		\midrule
		Encoding & 60s & 60ms$^\dag$ & 2 mins & {\bf 40ms} & {\bf 16ms} \\
		Render & 58ms & 31ms & 24ms & 24ms & 24ms \\
		\bottomrule
    \end{tabular}
    \label{tab:runtime}
    \setlength{\tabcolsep}{6pt}
\end{table}

\section{Related work}
\label{sec:related_works}
Our work touches on light fields, few-shot view synthesis, learning with synthetic data, 3D-aware portrait generation, and GAN inversion.  Our focus is on real-time view synthesis from a single image, and we do not address portrait relighting or editing. Tab.~\ref{tab:runtime} summarizes runtime for inferring 3D representations from an input and rendering. Our one-shot method is three orders of magnitude faster than the NeRF-based state-of-the-art methods for inference, enabling a real-time pipeline. 

\paragraph{Light Fields and Image-Based Rendering.} 
View synthesis or image-based rendering has a long history in computer graphics and vision~\cite{ChenWilliams,McMillan}, and has often been framed in terms of reconstructing the light field~\cite{Levoy,Gortler}.  However, those methods typically required hundreds of views.  Subsequent light field approaches demonstrated few-shot general~\cite{Nima} and even single image view synthesis for categories~\cite{PratulICCV}, but required light field camera training data.  More recently, neural-field based approaches~\cite{mildenhall2020nerf,Yiheng2022neuralfield} combine recent neural implicit 3D representations~\cite{park2019deepsdf,mescheder2019occupancy,chen2019learning,sitzmann2019srns} with volume rendering for novel view synthesis, but require a large number of input images per scene.  

\paragraph{Few-shot novel view synthesis.}
Some recent work extends NeRF for training from even a single view~\cite{Xu_2022_SinNeRF} or for few-shot novel view synthesis using fully implicit 3D representations~\cite{li2022symmnerf,yu2021pixelnerf,grf2020,jang2021codenerf}, 3D convolutions~\cite{mvsnerf,xianggang2022pvserf}, or Transformers~\cite{lin2023visionnerf, wang2021ibrnet}. However these approaches do not generate novel views in real-time. Moreover, all of the above approaches need multi-view images to train their models. Our method, on the other hand, only needs synthetic images generated from a pre-trained 3D GAN, which is trained by a collection of single-view images. FWD~\cite{Cao2022FWD} builds on top of SynSin\cite{wiles2020synsin} for real-time novel view synthesis using depth-based image warping, but requires depth data from multiview stereo or a depth sensor. Yet another family of approaches is the geometry-free method\cite{sajjadi2022scene,rombach2021geometryfree, ren2022look}, but they need a large number of images to learn precise ray transformations; otherwise it may lead to blurry or multiview inconsistent results.

\paragraph{Learning with synthetic data.}
Synthetic data provides useful supervision for training a deep learning model when ground truth data is not available. Previous methods used synthetic data for various deep learning-based tasks such as dense visual alignment~\cite{peebles2022gansupervised}, 3D face reconstruction~\cite{pan2020gan2shape,wood2022dense} and analysis~\cite{Wood_2021_fakeit}, portrait normalization~\cite{zhang2020portrait,nagano2019deepfacenorm}, and semantic segmentation~\cite{zhang21datasetgan,Tritrong2021RepurposeGANs}. Some previous work used synthetic face portrait images generated by rendering 3D face assets using a phycally-based pathtracer to train a model for portrait relighting ~\cite{yeh2022learning} or relighting and view synthesis~\cite{sun2021nelf}. %
Since the CG rendering exhibits a synthetic look, they need an additional step to adapt to real images. %
Other concurrent work~\cite{ko20233d3dganinversion} uses a discrete number of pre-generated synthetic images from a 3D-aware generator~\cite{eg3d2022} for 3D GAN inversion tasks. Instead, we generate an unlimited amount of synthetic data in the training loop and show that on-the-fly camera augmentation is critical for generalization to real images for synthetic data training.

\paragraph{3D-aware portrait generation and manipulation.}
For a well-known category of object, such as human faces, previous work~\cite{Khakhulin2022ROME,Gao-portraitnerf,athar2022rignerf,hong2021headnerf,Mihajlovic:ECCV2022,wang2022morf,abs-1802-05384,nagano2018pagan,kim2018deep} used 3D face priors for few-shot portrait synthesis. While the face priors provide additional capabilities for facial manipulations and expression retargeting~\cite{seol2011retargeting}, they don't generalize beyond humans. 
Recently, 3D aware-image generation approaches~\cite{hologan,graf,chan2020pi,Niemeyer2020GIRAFFE} started to tackle the problem of unconditional generation of photorealistic 3D representations from a collection of single-view 2D images. By combining neural volumetric rendering\cite{mildenhall2020nerf} and generative adversarial networks (GANs)\cite{goodfellow2014generative}, recent 3D GAN approaches\cite{eg3d2022,orel2022stylesdf,gu2021stylenerf,deng2022gram,xiang2022gramhd,epigraf,Zhou2021CIPS3D,zhang2022mvcgan,xu2021volumegan,rebain2022lolnerf} started to demonstrate an ability to generate high-resolution multi-view consistent images and geometry of a category of objects. 
We adapt the efficient triplane 3D representation from EG3D~\cite{eg3d2022} and demonstrate single-view novel view synthesis on similar categories.  

\paragraph{3D GAN inversion.}
Following the success of GAN inversion in 2D domains for image editing and manipulations\cite{richardson2021encoding, tov2021e4e, dinh2021hyperinverter,alaluf2021restyle,wang2021HFGI}, existing 3D GAN inversion methods~\cite{ko20233d3dganinversion,sun2022ide,lin20223dganinversion} project a given image to variants of the pre-trained StyleGAN2 latent space~\cite{Karras2020stylegan2,abdal2019image2stylegan}. \revision{Assuming multiview images, FreeStyleGan~\cite{leimkuhler} proposes to map projected camera coordinates to a subject-specific StyleGAN2 latent space which allows the subject to be rendered from specified cameras under the constraints of the StyleGAN prior.} While this global latent space provides an additional ability for 3D-aware portrait \textit{editing}, the StyleGAN2 latent space trades off reconstruction fidelity for editability, making the exact reconstruction of the input image challenging. Thus, existing 3D GAN inversion approaches require an approximate camera pose and slight generator weight tuning~\cite{roich2021pivotal, Feng2022nearperfect} at test time to reconstruct out-of-domain input images. %
Our feed forward encoder takes an unposed image as input and does not need test-time optimization for camera poses unlike concurrent work~\cite{ko20233d3dganinversion}.

\revision{
\paragraph{Talking-head generators.}
Given a single target portrait and a driving video, recent talking-head generators can reenact the portrait by transferring facial expressions and head poses from the driver video~\cite{wang2022latent,hong2022depth,Zakharov20,Drobyshev22MP,wang2021facevid2vid,tps2022,doukas2020headgan}. Trained by video datasets, their methods mainly focuse on talking-head video generation by manipulating avatar poses and expressions within a 2D portrait. As such, they do not predict volumetric representations that allow free viewpoint rendering including background and do not provide dense 3D geometry like our method. Therefore, we do not compare to these approaches. 
}

\begin{figure*}
    \centering
    \includegraphics[width=\textwidth]{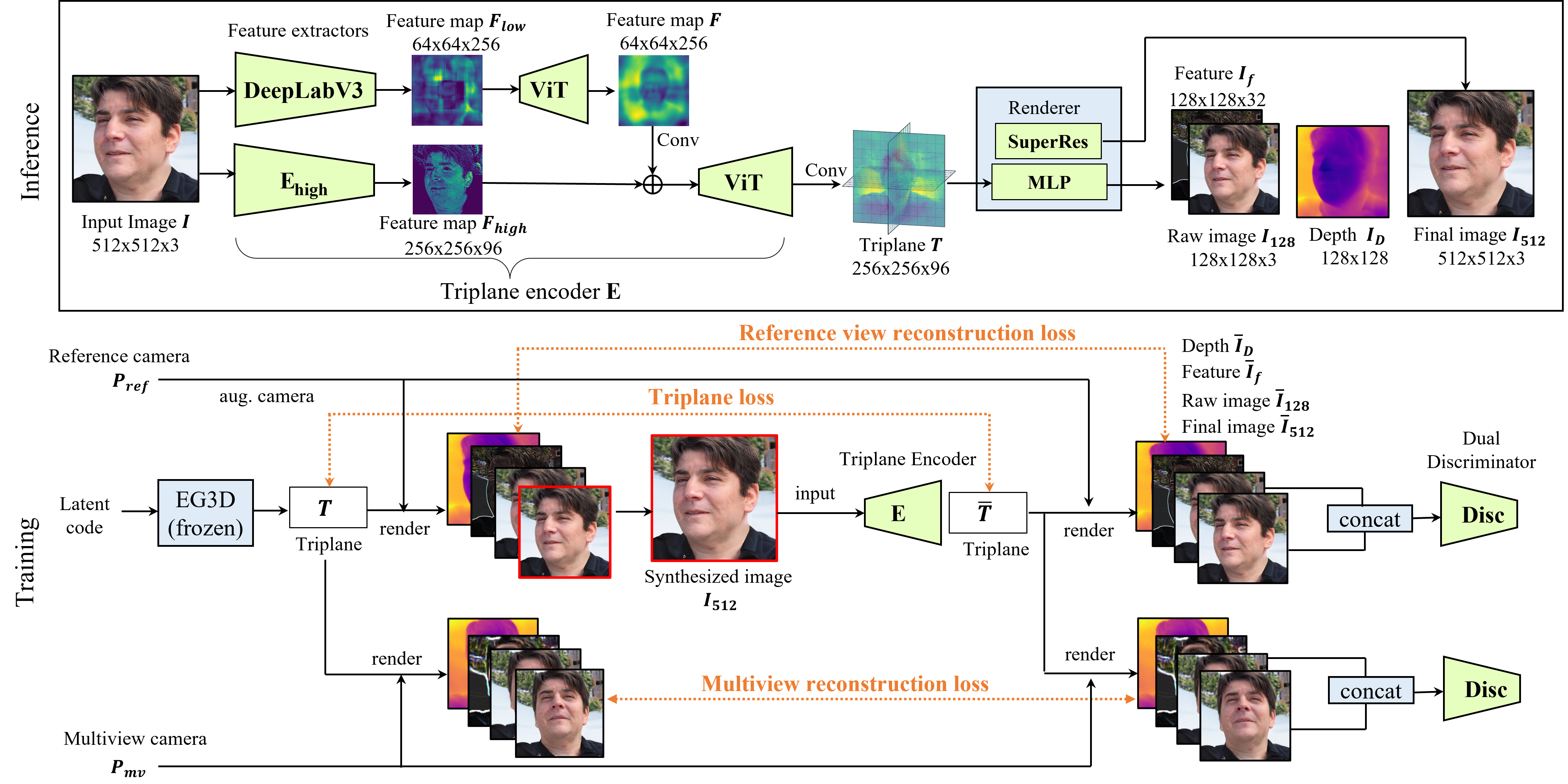}
    \caption{Inference and training outline for our pipeline. At inference, we take an unposed image, and extract low resolution features $\bm F_\text{low}$ with a DeepLabv3 backbone. These features are fed to a ViT yielding $\bm F$ and then concatenated after convolution with high-resolution features $\bm F_\text{high}$ before being decoded with a ViT and convolutions to a triplane representation. These features condition the volumetric rendering process which yields depth, feature, color, and superresolved images. During training, we sample an identity from EG3D and then render two supervision views. The first serves as the input to our encoder, which predicts a triplane, which conditions volume rendering from the same two views. The rendering results are compared with those of EG3D as outlined in Sec. \ref{sec:method}. Feature maps are visualized for illustration.}
    \label{fig:pipeline}
\end{figure*}

\section{Preliminaries: Triplane-based 3D GAN}
\label{sec:prelims}
We first give an overview of the state-of-the-art 3D GAN method, EG3D, \cite{eg3d2022} from which our method will distill knowledge. EG3D learns unconditional 3D-aware image generation from a collection of single-view images and corresponding noisy camera poses, where each image has resolution $512\times512$. As mentioned in Sec.~\ref{sec:related_works}, EG3D makes use of a hybrid triplane representation to condition the neural volumetric rendering process, whereby three 2D feature grids are stored along each of the three canonical planes--$xy$, $xz$, $yz$. Using a StyleGAN2 generator~\cite{Karras2020stylegan2}, EG3D maps a noise vector and conditioning camera poses to a triplane representation $\bm T\in\mathbb{R}^{256\times 256\times 96}$ which corresponds to the $3$ axis-aligned planes, each with $32$ channels. These features condition the neural volumetric rendering.

To assign a point $\bm x\in\mathbb{R}^3$ with its feature, 
color and volume density, $(\mathbf{f}, \mathbf{c},\sigma)$, a lightweight MLP decodes the three feature vectors gathered by projecting $\bm x$ to each of the canonical planes:
\begin{equation}
(\mathbf{f}, \mathbf{c},\sigma) = \text{MLP}( \Phi (\bm f_{xy}, \bm f_{xz}, \bm f_{yz}) ),
\label{eq:1}
\end{equation}
where $\bm f_{ij}$ are the features gathered by projecting $\bm x$ to the $ij$ plane and bilinearly interpolating the nearby features, and $\Phi$ is the mean operator. Note that output values including the color are independent of viewing direction and only depend on $\bm x$.  
By accumulating many points along rays, and performing volume rendering~\cite{Max:1995} as in NeRF \cite{mildenhall2020nerf}, one may render a feature image $\bm I_f\in\mathbb{R}^{32 \times 128 \times 128 }$ and a raw neural rendering RGB image $\bm I_{128} \in\mathbb{R}^{3 \times 128 \times 128 }$ from a given camera pose. In practice, $\bm I_{128}$ corresponds to the first three channels of the feature image  $\bm I_f$. %

We additionally extract a dense depth map $\bm I_D\in\mathbb{R}^{128 \times 128 }$ from this volume rendering, which we use later to supervise our model.  
The neural rendered images $\bm I_{128}$ and $\bm I_f$ are then
fed to a 2D superresolution network, which yields the final superresolved rendering output: 
\begin{equation}
\text{SuperRes}(\bm I_f, \bm I_{128}) = \bm I_{512} \in \mathbb{R}^{3 \times 512 \times 512 }. 
\label{eq:2}
\end{equation}
This 3D GAN pipeline is trained end-to-end following 2D GAN training with a 2D (dual) discriminator. The reader is referred to the original paper~\cite{eg3d2022} for full details. %

The efficient design of EG3D allows rendering from a triplane at 42 fps on the RTX 3090.  %
At the same time, EG3D provides comparable quality to even the state-of-the-art 2D GANs by FID. These attributes provide a strong basis for supervising our encoder-based method using EG3D-generated synthetic data.

\section{Method}
\label{sec:method}
\begin{figure*}[h!]
    \centering
    \includegraphics[width=\linewidth]{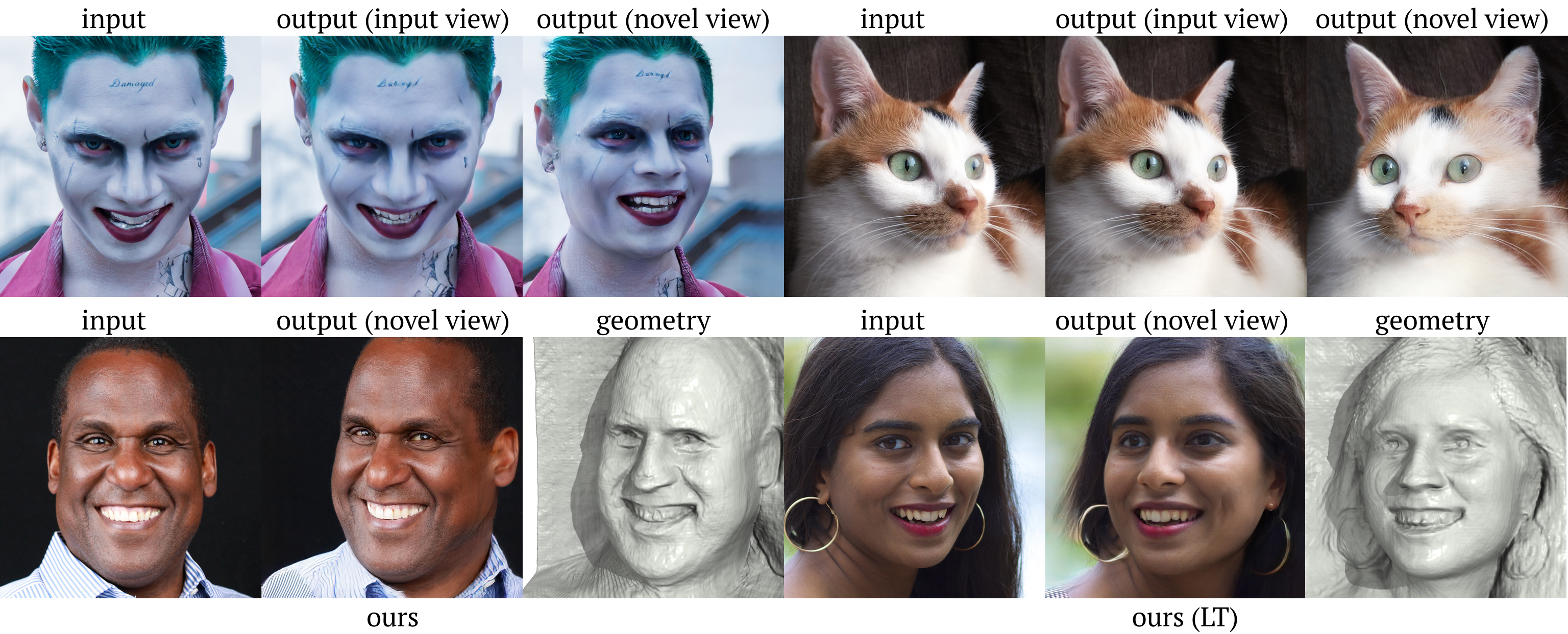}
    \caption{FFHQ and AFHQ qualitative results from our model (left) and our lightweight model (LT) (right). We showcase reconstructed input and novel views, or the learned geometry. In the bottom-right, note our model's ability to infer structure consistent with the input behind occlusion. %
    Credits to YuChen Cheng, Montclair Film, Lydia Liu.}
    \label{fig:results}
\end{figure*}

Our goal is to distill the knowledge of a fully trained EG3D generative model 
(learned over a
category or set of categories) into a feedforward encoder
pipeline that can {\em directly map an unposed image to a canonical triplane
3D representation\footnote{
Note that each
category has a different notion of canonical representation: for human
faces, the center of the head is the origin, and planes orthogonally
intersect the head up-to-down, left-to-right, and front-to-back.
} which can be decoded with a NeRF.}
This pipeline requires only a single feedforward network pass, thus avoiding
the expensive GAN inversion process, while allowing free
viewpoint re-rendering of the input in real-time.  
Note that our contribution 
focuses on the image-to-triplane encoder and associated synthetic training method, as shown in the pipeline of Fig.~\ref{fig:pipeline}.  We make use of the MLP volume renderer and superresolution architectures from EG3D as per Eqns~\ref{eq:1} and~\ref{eq:2}  and train all the components end-to-end.  In Tab.~\ref{tab:runtime}, the top row shows that our image to triplane inference runs at up to 60 fps (16 ms), %
while rendering has identical performance to EG3D (bottom row of Tab.~\ref{tab:runtime}).

\subsection{Triplane encoder}
We note that inferring a canonicalized 3D reprentation (i.e., the inferred 3D representation is frontalized and aligned) from an arbitrary RGB image while simultaneously synthesizing precise subject-specific details from the input is a highly non-trivial task. We break this challenge into the two-fold goals: 1) to create a canonicalized 3D representation of the subject from an image, and 2) to render high-frequency person-specific details. We note these goals are often at odds with one another, and exemplify the bias-variance tradeoff whereby the output will resemble the input well, but may not be correctly canonicalized in 3D (see Fig. \ref{fig:augmentation_ablation}), or the output will have the correct 3D structure, but not resemble the 2D input image (see Fig.~\ref{fig:transformer_ablation}). %
Our encoder manages to accomplish both of these goals simultaneously. Specifically, we develop and train a hybrid convolutional-Transformer encoder, $\mathbf{E}$, which maps from an unposed RGB image, $\bm I$, to the \emph{canonical} triplane representation. %

As seen in the upper half of Fig.~\ref{fig:pipeline}, the architecture of our encoder begins with a fast convolutional backbone, DeepLabV3 \cite{chen2017rethinking}, which extracts robust low-resolution features, $\bm F_\text{low} = \text{DeepLabV3}(\bm I)$. These features are then fed to a Vision Transformer (and CNN) which gives a global inductive bias to the intermediate output features, 

\newcommand{\Conv}{\mathrm{Conv}}
\newcommand{\Vit}{\mathrm{ViT}}
\begin{equation}
\bm F = \Conv(\Vit(\bm F_\text{low})),
\label{eq:3}
\end{equation}
where $\Conv$ is a CNN and $\Vit$ is the Vision Transformer Block from Segformer \cite{xie2021segformer} with efficient self-attention. \revision{We choose the Segformer ViT for two reasons: 1) it was designed to quickly map to a high-resolution output space similar to a triplane, and 2) the efficient self-attention mechanism allows the use of high-resolution intermediate feature maps so that all information flows from input to triplane. }

We consider the ViT features as having successfully created a canonicalized 3D representation of the subject (completing the step 1 above), and found during our experimentation that this shallow encoder is sufficient to reasonably canonicalize a subject, yet cannot represent important high-frequency or subject-specific details like strands of hair or birthmarks. %

In order to simultaneously complete the second step (adding high-frequency detail), we next reincorporate high-resolution image features. %
We convolutionally encode the image again with only a single downsampling stage with encoder $\mathbf{E}_\text{high}$ to obtain features $\bm F_\text{high} = \mathbf{E}_{\text{high}}(\bm I)$. These are concatenated with the extracted global features and passed through another Vision Transformer, which is finally decoded to a triplane with convolutions as seen in Fig.~\ref{fig:pipeline}. Thus, the output of our encoder has the following form:
\begin{equation}
\bm T = \mathbf{E}(\bm I) = \Conv(\Vit(\bm F \oplus \bm F_{\text{high}})),
\label{eq:4}
\end{equation}
where $\oplus$ denotes concatenation along the channel axis, and $\bm T$ is triplane feature representation used in Sec.~\ref{sec:prelims}.

\subsection{Training with synthetic data}
\newcommand{\mv}{\mathrm{mv}}
\newcommand{\refe}{\mathrm{ref}}

As seen in Fig.~\ref{fig:pipeline} in the training step, we train our triplane encoder with synthetic data. Sampling a latent vector and passing it through the EG3D generator yields a corresponding triplane, $\bm T$. Given camera parameters $\bm P$ (a focal length, principal point, camera orientation and position), we can render any image from the frozen EG3D generator and $\bm T$. At each gradient step, we synthesize two images of the same identity (same latent code) from a reference (input) camera $\bm P_{\refe}$ and another camera $\bm P_{\mv}$ for multiview supervision. Using the same notation as in Sec.~\ref{sec:prelims}, each rendering pass will give us four images: $\bm I_f$, $\bm I_{128}$, $\bm I_{512}$, and $\bm I_D$ as seen in Fig.~\ref{fig:pipeline}. 

Again as shown in Fig.~\ref{fig:pipeline}, the input to our encoder is the high-resolution image $\bm I_{512}$ (highlighted in red) rendered from the input camera $ \bm P_\refe$, so that $\overline {\bm T}=\mathbf{E}(\bm I_{512})$. We then use $\overline{\bm T}$ to condition the volume rendering process from both camera $\bm P_\refe$ and $\bm P_\mv$, to get two more sets of four images, which we denote as $ \overline{\bm I}_f$, $\overline{\bm I}_{128}$, $\overline{\bm I}_{512}$, and $\overline{\bm I}_D$. Our loss intuitively compares those quantities synthesized by EG3D and those created by our encoder, along with a generative adversarial objective as follows: %
\begin{eqnarray}
L =\ L_\text{tri}+ L_\text{col} + L_{\text{LPIPS}} + L_\text{feat} + \lambda_{1} L_\text{adv} + \lambda_{2} L_\text{cate}
\label{eq:5}
\end{eqnarray}
$L_\text{tri}$ is the L1 loss between $\bm T$ and $\overline {\bm T}$; $L_\text{col}$ is the mean L1 loss computed between both sets of pairs $(\bm I_{128}, \overline{\bm I}_{128})$ and $(\bm I_{512}, \overline{\bm I}_{512})$; $L_\text{LPIPS}$ is the LPIPS perceptual loss~\cite{zhang2018perceptual} computed over both sets of pairs $(\bm I_{128}, \overline{\bm I}_{128})$ and $(\bm I_{512}, \overline{\bm I}_{512})$; $L_\text{feat}$ is the mean L1 loss computed between the pairs $(\bm I_{f}, \overline{\bm I}_{f})$;
$L_\text{adv}$ is the adversarial loss using a pretrained dual discriminator from EG3D which is fine-tuned during training; $\lambda_{1}$ is $0.1$ for the reference image or $0.025$ for the multiview image; and $L_\text{cate}$ is an optional category-specific loss. For human faces, we use $\lambda_{2}$ to be $1$ with face identity features from ArcFace ~\cite{deng2018arcface} following practice in 2D GAN inversion~\cite{richardson2021encoding,tov2021e4e}. For cat faces, we set $\lambda_{2}$ to $0$. 
This objective is optimized end-to-end, i.e., with respect to all of the parameters of the encoder, rendering and upsampling modules. \revision{Note that the rendering, upsampling, and dual discriminator modules are all fine-tuned from the pretrained EG3D.} However, the dual discriminator in our pipeline doesn't rely on \emph{any} real data; instead, we train this discriminator to differentiate between images rendered from our encoder model and images rendered from the frozen EG3D. \revision{An ablation showing its effectiveness is provided in Tab.~\ref{tab:disc_ablation} and Fig.~\ref{fig:disc_ablation}.} %

\paragraph{On-the-fly augmentation.}
Naively optimizing this objective will yield a model which performs almost perfectly on synthetic data, but lacks the ability to generalize to real images (see Fig.~\ref{fig:augmentation_ablation}). In order to remedy this, we augment the standard EG3D rendering method which assumes a fixed camera roll, focal length, principal point and distance from subject. In contrast, we sample all four of these values from random distributions to choose the camera parameters $\bm P_{\mathrm{ref}}$. The details of these distributions for each dataset are given in the supplement. For $\bm P_\mathrm{mv}$, we choose fixed values as in the EG3D model. For $\bm P_{\mathrm{ref}}$, we sample the cameras from a pitch range of $\pm 26^{\circ}$ and yaw range of $\pm 49^{\circ}$ relative to the front of a human face. For $\bm P_\mathrm{mv}$, we sample the cameras from a pitch range of $\pm 26^{\circ}$ and yaw range of $\pm 36^{\circ}$ relative to the front of a human face. This allows the supervision of our model to happen with highly variable camera poses, forcing the model to learn to effectively canonicalize and infer from challenging images as seen in Fig.~\ref{fig:results}. %

\paragraph{Implementation details}
Before training with the full adversarial objectives in Eqn.~\ref{eq:5}, we warm up the model by training over 30k iterations without the adversarial loss and continue to train the model with the full loss functions in Eqn.~\ref{eq:5} over ~220k iterations. Since we sample two camera poses per iteration (with batch size 32), we effectively use over 16 million images during the training, which is not obtainable from real images (nor even physically-based rendered images) in practice. For full implementation details, please refer to the supplement.
We train two encoders with two different compute budgets: "Ours", %
which has 87M parameters and "Ours (LT)", a lightweight model (LT) which has 63M parameters. The main difference between the two is in resolution of the intermediate feature maps, which result in fewer parameters in the LT model, but both contain the same structure outlined above.
"Ours" runs in 22ms on a single A100 GPU (where rendering takes 15ms) and 40ms on RTX 3090 as seen in Table \ref{tab:runtime}. "Ours (LT)" runs in just 16ms on RTX 3090, while retaining strong performance (see the numerical evaluations in Tabs.~\ref{tab:numerical_results} and \ref{tab:depth}). Figures~\ref{fig:teaser} and \ref{fig:results} show the qualitative outputs from both models.

\section{Results}
\label{sec:experiments}

\begin{figure*}[!ht]
    \centering
    \includegraphics[width=0.94\textwidth]{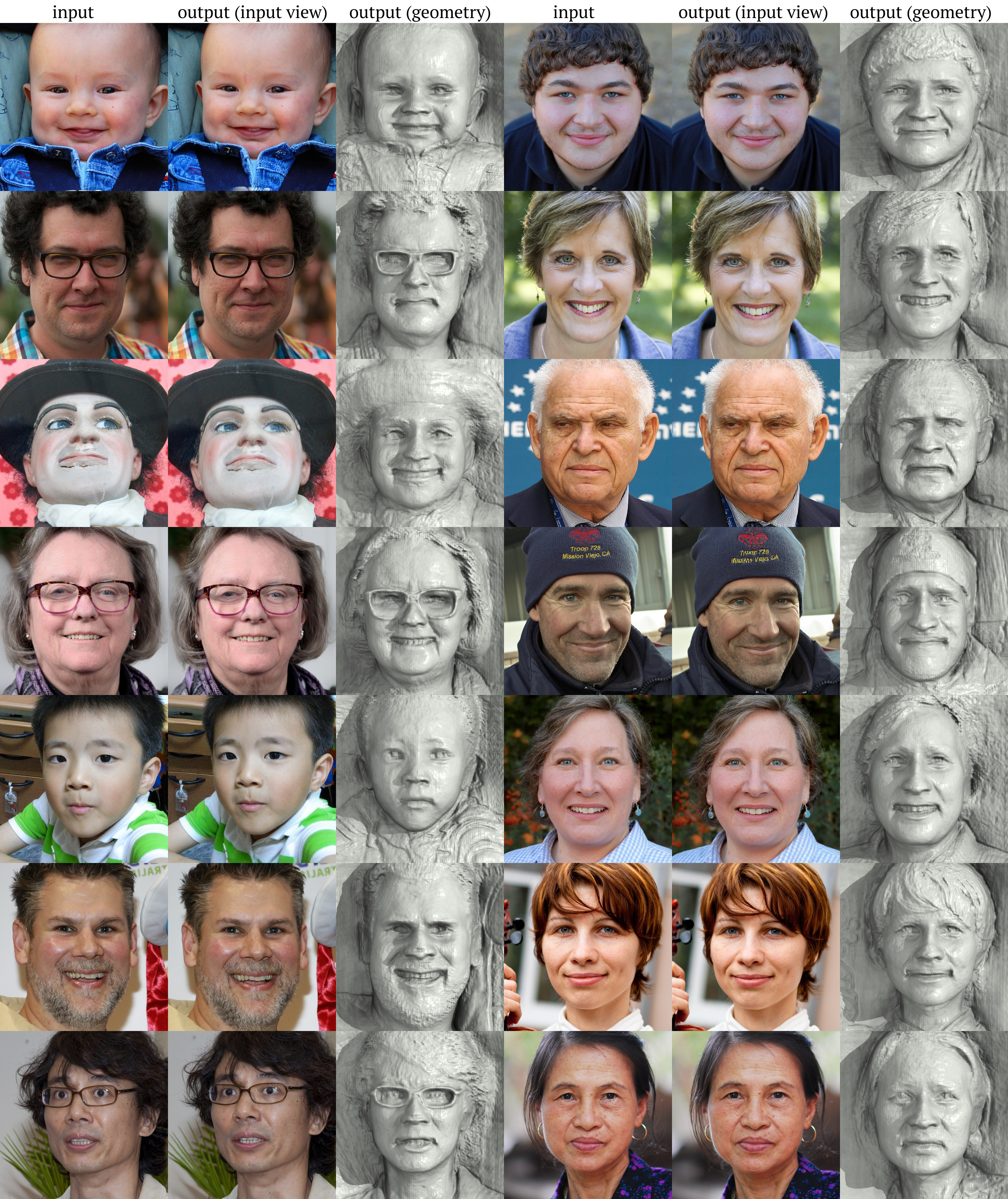}
    \caption{Qualitative results displaying our model's reconstruction on the input view, and the learned geometry from the frontal view. The reconstructed geometry remains faithful to the input image. Credits to Devon Weller, Jamie, SupportPDX, Mary Sawatzky, map, Herzliya Conference, Helse Midt-Norge, Tom Munnecke, pter tr, UGA CAES/Extension, Rare Cancers Australia, Vladimir Agafonkin, Michael E. Macmillan, Nguyen Hung Vu.}
    \label{fig:geo2}
\end{figure*}

\begin{figure*}
    \centering
    \includegraphics[width=\textwidth]{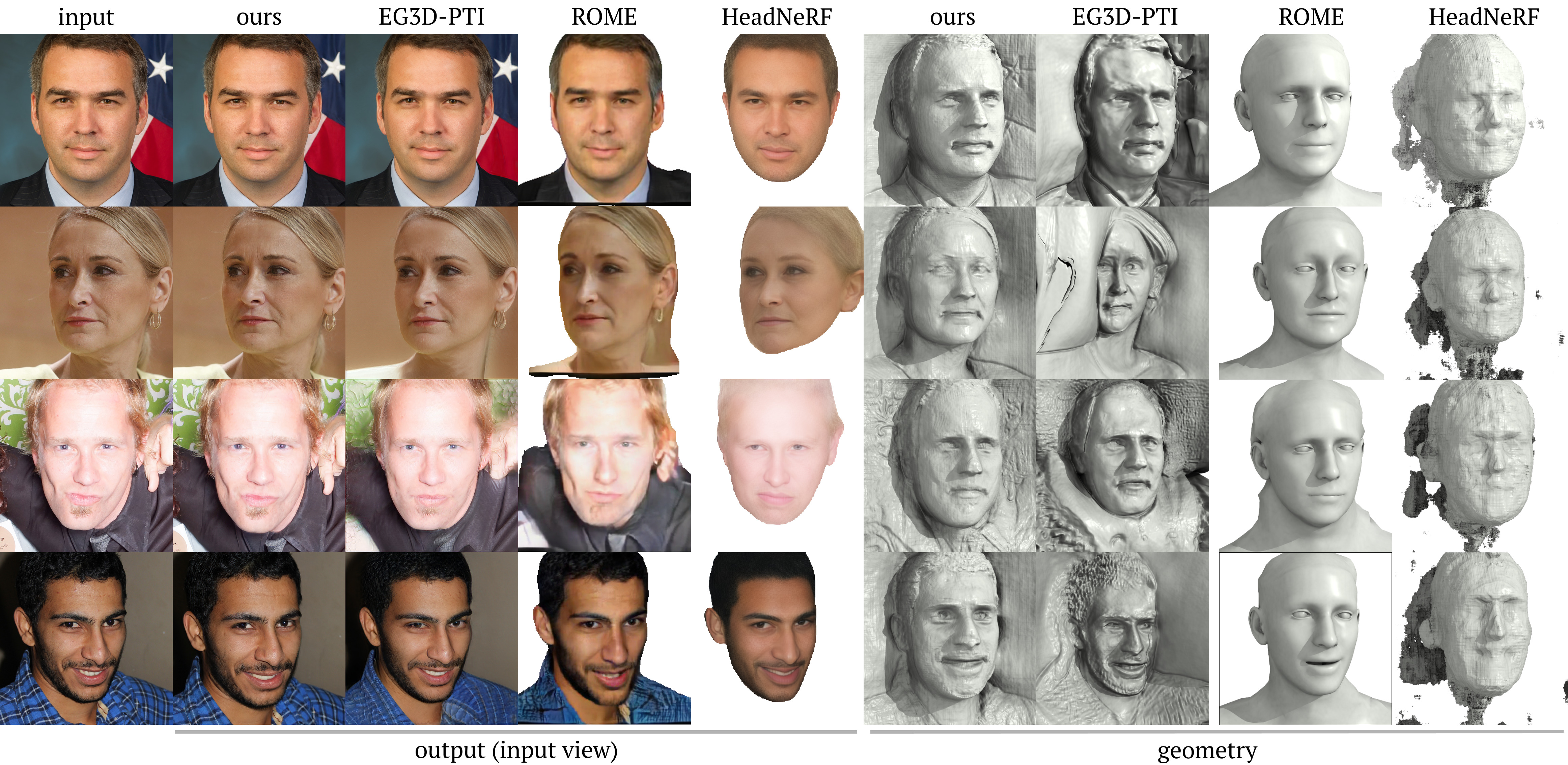}
    \caption{Qualitative results displaying our model in comparison to baseline methods HeadNeRF, ROME, and EG3D-PTI, comparing the image quality (left) and reconstructed geometry (right). EG3D-PTI occasionally exhibits corrupted 3D geometry (2nd and 4th rows) when the input is side view, indicating that the learned 3D prior alone is not enough to ensure robust reconstruction. Credit to U.S. Dept. of HUD, Cristina Cifuentes, Rainforest Action Network, CENA MINEIRA.}
    \label{fig:comparison_1}
\end{figure*}

\begin{figure}
    \centering
    \includegraphics[width=\linewidth]{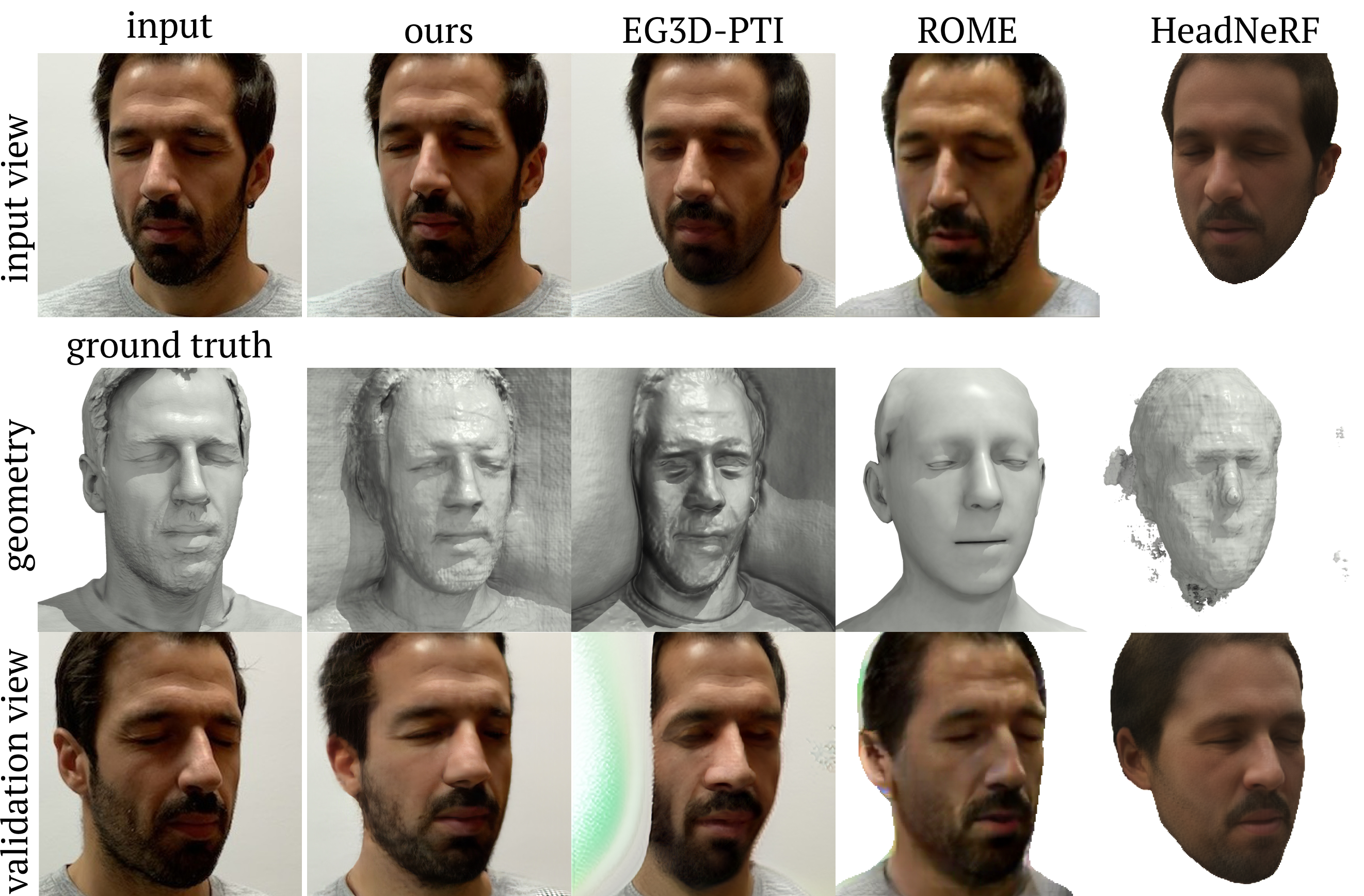}
    \caption{Ground truth comparisons on the H3DS dataset including ground truth geometry (second row) and unseen validation view (third row). Since the H3DS ground truth data has inconsistent lighting, the lighting discrepancy is expected for the validation view.}
    \label{fig:H3DS}
\end{figure}

\begin{figure*}
    \centering
    \includegraphics[width=\textwidth]{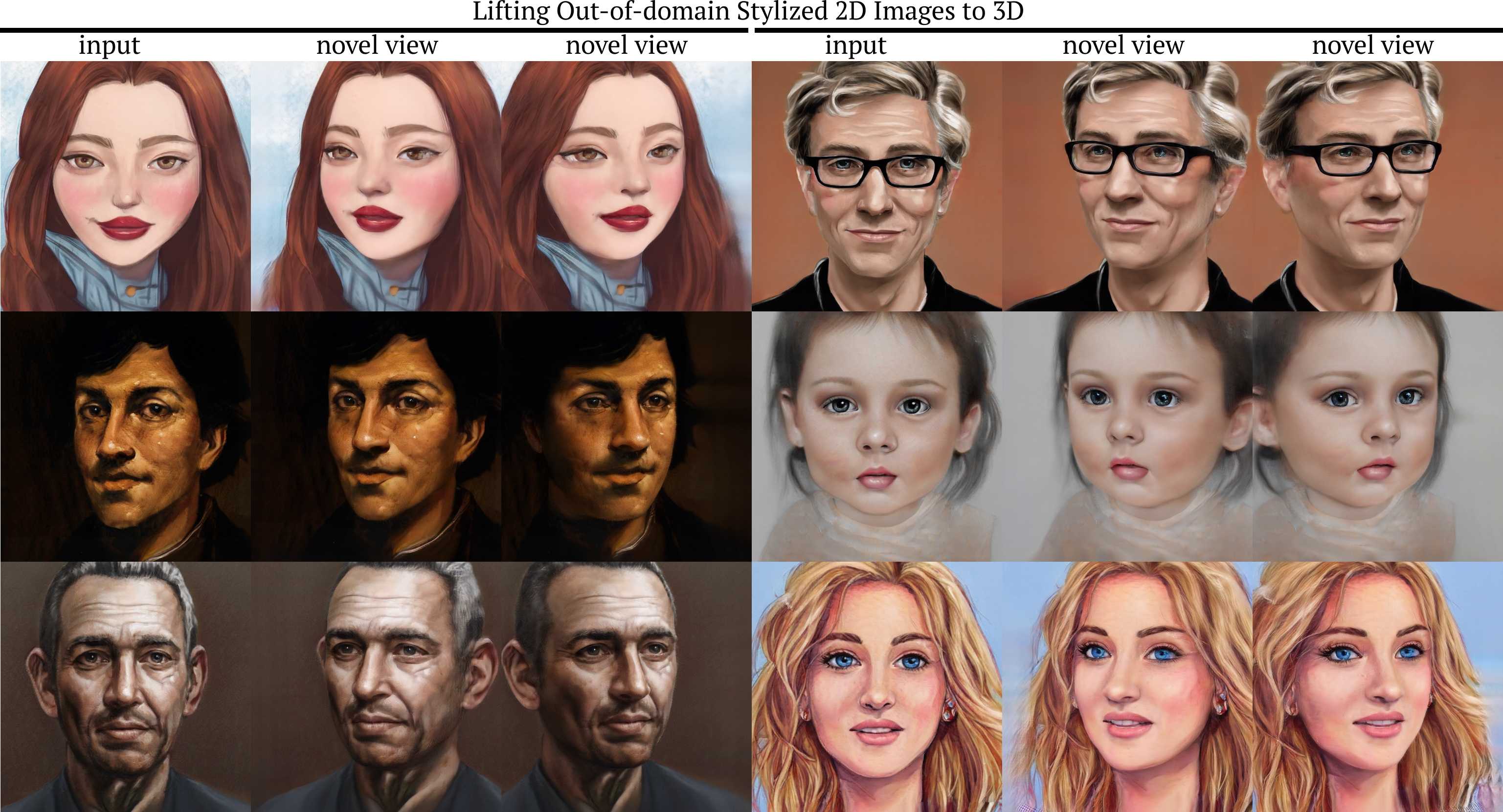}
    \caption{Qualitative results displaying our model's ability to lift StyleGAN2-generated drawings and paintings to 3D. These results display the generalizability of our model, as it can canonicalize out-of-domain drawings and portraits, lifting them to 3D. }
    \label{fig:stylized}
\end{figure*}

\begin{figure*}[!ht]
    \centering
    \includegraphics[width=\linewidth]{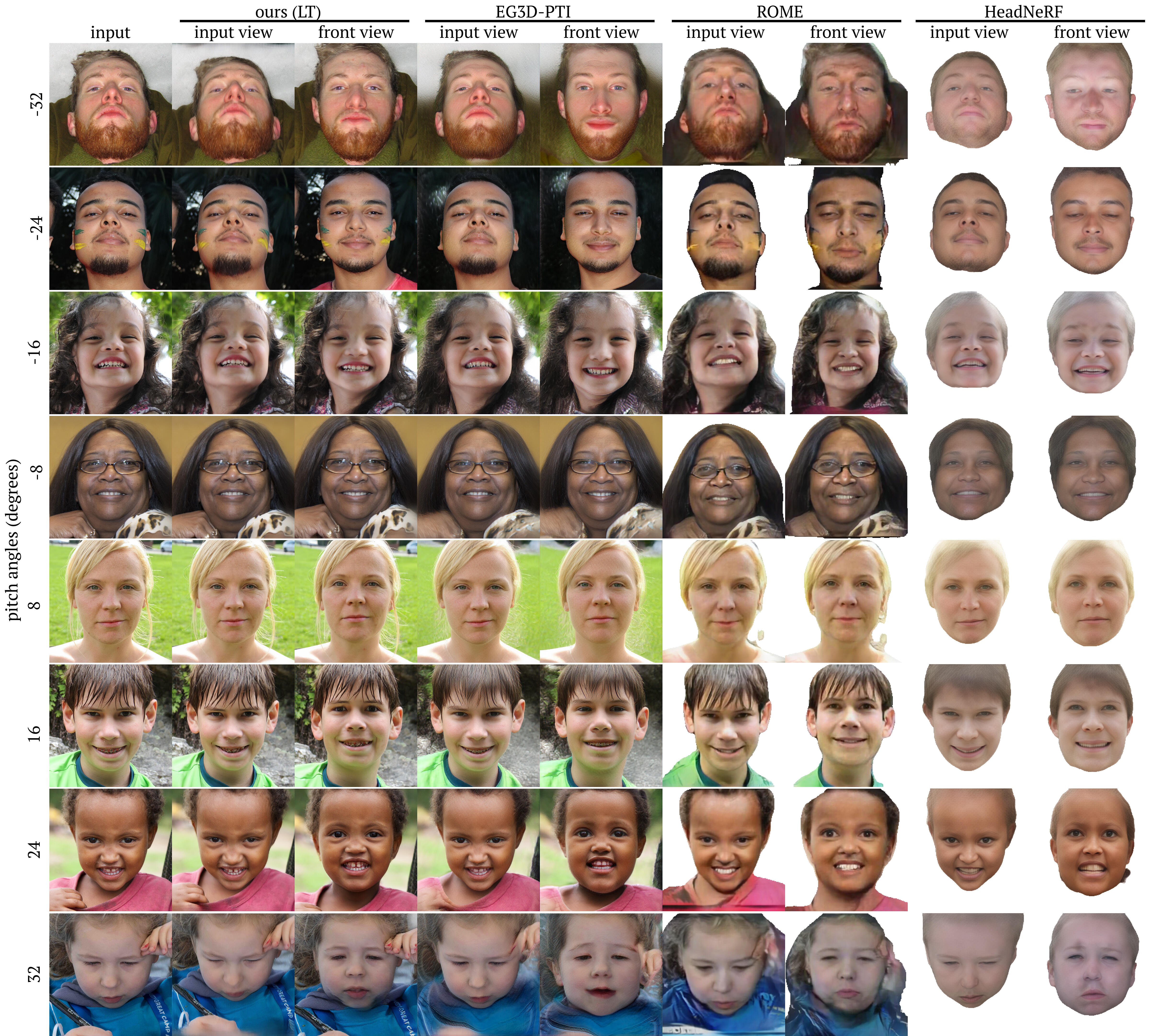}
    \caption{Comparison to baselines at various input pitch angles. Credits to Bjørnar Tollaksen, Juliana Martuscelli, 
The 621st Contingency Response Wing, U.S. Army Security Assistance Command, Sam Wadman, Laity Lodge Family Camp, U.S. President's Malaria Initiative, SickKids Foundation. }
    \label{fig:pitch}
\end{figure*}

 \begin{figure*}[!ht]
    \centering
    \includegraphics[width=\linewidth]{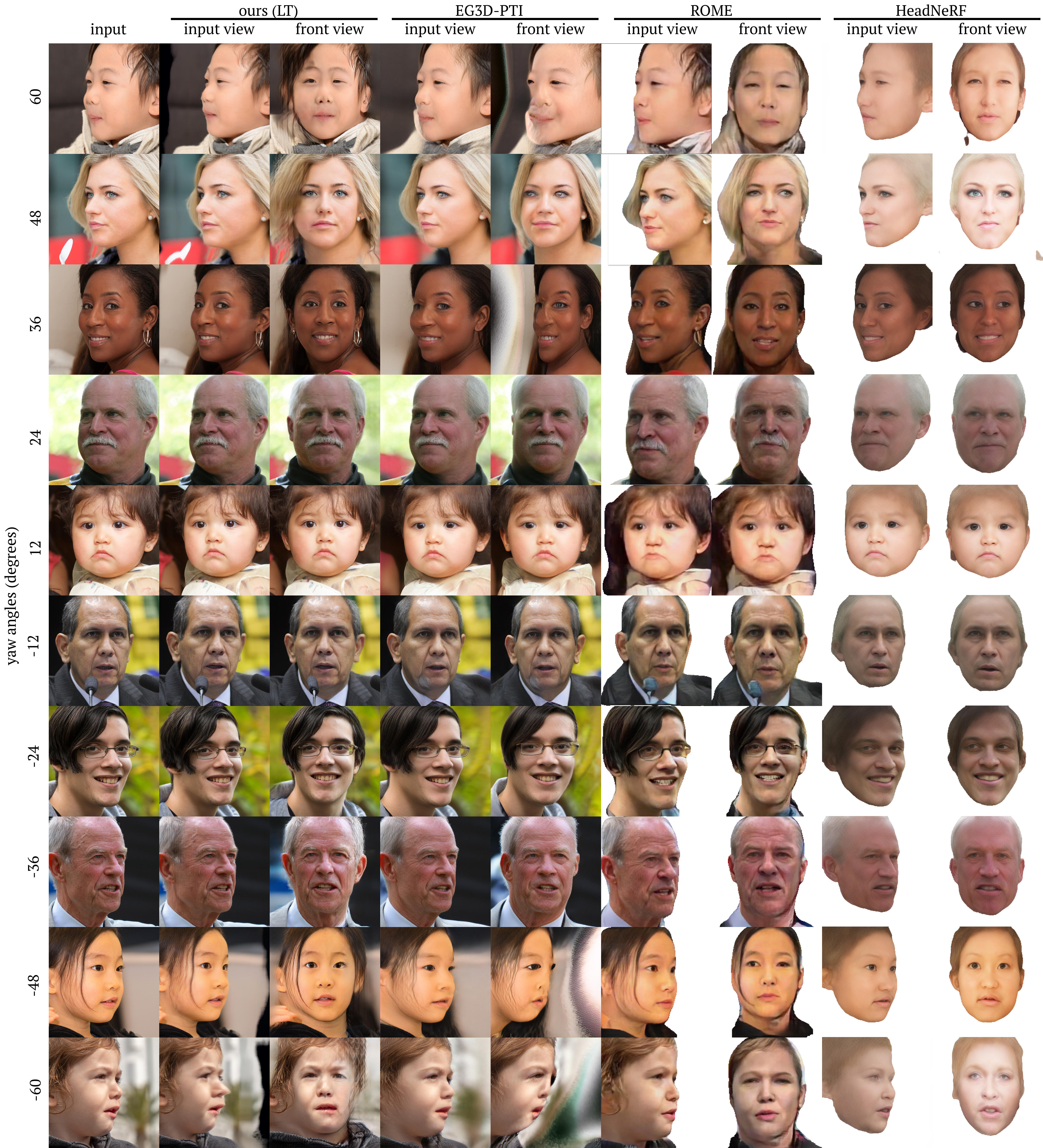}
    \caption{Comparison to baselines at various input yaw angles. Credit to Lionel AZRIA, justinkim1, nonorganical, Paradox Wolf, Agência Senado, Ademir Brito, Ariana Vincent, Jay Weenig, John Benson, Seong Bae.}
    \label{fig:yaw}
\end{figure*}

We evaluate methods for single-view novel view synthesis on 3 main aspects (1) 2D image reconstruction (LPIPS~\cite{zhang2018perceptual}, DISTS~\cite{ding2020iqa}, SSIM~\cite{wang2004ssim}) and likeness (identity consistency) (2) general image quality (FID~\cite{heusel2017fid}) and (3) 3D reconstruction quality (depth, and pose estimation). 
For the reconstruction tasks, we need to re-render our outputs to the input views for the purpose of the evaluation using a camera pose estimated using an off-the-shelf pose predictor~\cite{deng2019accurate}. However, we noticed that errors present in the estimated poses create a small image misalignment between the ground truth and our feedforward results (as opposed to inversion models which directly optimize for the given view), making the raw pixel metrics like PSNR and SSIM unreliable.   
For this reason, we mainly rely on the deep perceptual image metrics such as LPIPS and DISTS, which judge that the given images are of the same perceptual quality for our evaluation.  
Nonetheless, we report SSIM results in the main paper and include PSNR results in the supplement along with an analysis of alignment issues. In the end, our experiments qualitatively and quantitatively support that our method achieves the state-of-the-art results on in-the-wild portraits as well as multiview 3D scan datasets. For more results, please refer to the supplement video.

\paragraph{Datasets.}
Our method is evaluated on FFHQ~\cite{karras2019style}, a representative dataset for high-quality in-the-wild human portraits, H3DS~\cite{ramon2021h3d}, which has high resolution ground truth 3D scans and 360º images of 23 human heads with associated camera calibrations, and AFHQv2 Cats~\cite{choi2020starganv2,Karras2021}, a collection of high-resolution in-the-wild portraits of cats.

\subsection{Comparisons}
\paragraph{Baselines.}
We compare our methods against three state-of-the art methods for 3D aware-image generation from a single image: ROME~\cite{Khakhulin2022ROME}, HeadNeRF~\cite{hong2021headnerf}, and EG3D-PTI, which combines an unconditional EG3D generator~\cite{eg3d2022} and Pivotal Tuning Inversion (PTI)~\cite{roich2021pivotal}. We also compare with EG3D itself as an unconditional reference on FID. We additionally provide extensive evaluations on our lightweight model (LT), which is introduced in Sec.~\ref{sec:method}.

\paragraph{Qualitative results.}
Fig.~\ref{fig:teaser} shows our qualitative results on FFHQ and AFHQ. %
 Fig.~\ref{fig:results} and Fig.~\ref{fig:geo2} show selected examples from FFHQ, demonstrating high-quality novel views and 3D geometry reconstructed by our method from a single portrait.
Fig.~\ref{fig:comparison_1} provides a qualitative comparison against baselines. While HeadNeRF and ROME provide adequate shapes and images, they need image segmentation as a preprocess, and struggle with obtaining photorealistic results. Despite the 20 mins of fine tuning, EG3D-PTI does not ensure the reconstruction looks photorealistic when viewed from a non-input view (see Fig.~\ref{fig:pti_comparison}). In contrast, our method reconstructs the entire portrait with accurate photorealistic details. Fig.~\ref{fig:H3DS} provides comparisons to the ground truth validation view and 3D scan on H3DS. The synthesized image and 3D geometry of ROME and HeadNeRF generally lack the fidelity and reconstruct only a part of the head. EG3D-PTI occasionally outputs a degenerate 3D shape due to the highly unconstrained nature of single-view training of the NeRF representation (see Figs~\ref{fig:teaser},~\ref{fig:comparison_1} and~\ref{fig:H3DS}). Our geometry retains overall the 3D shape as well as person-specific facial details. We also provide results on lifting 2D drawings and paintings into 3D in Fig.~\ref{fig:stylized}. While our method is never trained with stylized images, it can reasonably well handle those out-of-domain input images. \revision{Finally, we also show the outputs of our method in comparison to baselines at varying pitch and yaw in Figs~\ref{fig:pitch} and ~\ref{fig:yaw}, displaying the benefit of our method for photorealistic facial frontalization of challenging images. In comparison to baselines, our method's geometry does not collapse for challenging yaws as EG3D-PTI, and shows a significantly higher degree of photorealism than ROME and HeadNeRF.}%

\begin{table}[t]
    \centering
		\small
		\caption{Quantitative evaluation using LPIPS, DISTS, SSIM, pose accuracy (Pose) and identity consistency (ID) on 500 FFHQ images.  \textsuperscript{$\dag$}Evaluated only using the foreground on $256^2$ images.  \textsuperscript{$\ddag$}Evaluated only using the face region.}
    \begin{tabular}{@{\hskip 1mm}l c c c c c@{\hskip 1mm}}
				\toprule
		        	            & LPIPS $\!\downarrow$   & DISTS $\!\downarrow$   & SSIM $\!\uparrow$       & Pose $\!\downarrow$   & ID $\!\uparrow$  \\
        \midrule
        \midrule
        HeadNeRF$^\ddag$	    & .2502 		            & .2427                 & .7514              	& .0644                 & .2031 \\
        Ours$^\ddag$    	    & \textbf{.1240}            & \textbf{.0770}        & \textbf{.8246}	    & \textbf{.0490}        & \textbf{.5481}   \\
        \midrule
        ROME (256)$^\dag$ 	    & .1158 				    & .1058                 & .8257     	       & .0637			        & .3231 \\
        Ours$^\dag$     	    & \textbf{.0468}            & \textbf{.0407}        & \textbf{.8981}	    & \textbf{.0486}        & \textbf{.5410}   \\
        \midrule
        EG3D-PTI	            & .3236				    	& .1277                 & \textbf{.6722}		& .0575				    & .4650 \\
        Ours  			        & \textbf{.2692}            & \textbf{.0904}        & .6598                 & .0485                 & \textbf{.5426} \\
        Ours (LT)  			    & .2750                     & .1021                 & .6655                 & .\textbf{0448}        & .5404 \\
				\bottomrule
    \end{tabular}
    \label{tab:numerical_results}
\end{table}

\paragraph{Quantitative evaluations.}
Tab.~\ref{tab:numerical_results} shows numerical comparisons of our method against baselines on 500 randomly selected images from FFHQ. We measure the 2D image reconstruction quality in the input view using LPIPS, DISTS, and SSIM. We evaluate multiview consistency using poses (Pose) estimated from synthesized images by an off-the-shelf pose detector~\cite{deng2019accurate} following similar protocols as in previous work~\cite{eg3d2022,shi2021lifting}, and identity (ID) consistency by computing the mean of MagFace~\cite{meng2021magface} (not used in our training) cosine similarity scores between the input view and synthesized view from a random camera pose. Since HeadNeRF and ROME only produce the face region and the foreground respectively, we also provide the same metrics from our models evaluated on the same masked region.
Tab.~\ref{tab:numerical_results} shows that our model significantly outperforms the baselines on all the metrics except SSIM; our SSIM score is only marginally lower than EG3D-PTI despite the aforementioned issue of the image misalignment and the fact that EG3D-PTI directly optimizes the pixels for the evaluation view. The geometry evaluation in Tab.~\ref{tab:depth} on H3DS in which we compare the depths of the ground truth from the input view as predicted by each model validates that our models produce more accurate 3D geometry.

\begin{table}[t]
\setlength{\tabcolsep}{4.5pt}
    \centering
		\small
    \caption{Scale- and translation-invariant depth evaluation using ground truth geometry from H3DS datasets. \textsuperscript{\dag}Evaluated only using the face region.}
    \begin{tabular}{@{\hskip 1mm}l c c c c c@{\hskip 1mm}}
		\toprule
		Depth & H.NeRF & ROME & EG3D-PTI & Ours & Ours (LT)\\
		\midrule
		L1 $\!\downarrow$ & 0.108$^\dag$ & 0.054 & 0.071 & \textbf{0.048} & 0.049 \\
		RMSE $\!\downarrow$ & 0.147$^\dag$ & 0.084 & 0.101 & \textbf{0.074} & 0.075 \\
		\bottomrule

    \end{tabular}
    \label{tab:depth}
\setlength{\tabcolsep}{6pt}
\end{table}

\subsection{Ablation study}
We provide ablation studies comparing variants of our architecture and different training strategies. All variants are evaluated after training with 3M images. 
\paragraph{Inference time and number of parameters}
We compare the performance of two variants of our model, which have the same architecture but have different numbers of parameters and resolution of intermediate feature maps: %
"Ours" (87M params) and "Ours (LT)" (63M params). Tab.~\ref{tab:runtime} provide runtime comparisons of the two. Tabs.~\ref{tab:numerical_results} and ~\ref{tab:depth} provide several comparisons of the two on image reconstruction, the accuracy of 3D shapes and identity consistencies. These extensive evaluations suggest that our lightweight model retains very close performance to our full model despite running significantly faster. %
Figs.~\ref{fig:teaser} and ~\ref{fig:results} show qualitative samples from both our full model and our lightweight model. 

\paragraph{Effects of Transformers}
Fig.~\ref{fig:transformer_ablation} compares results obtained with or without the proposed Transformer layers in the encoder. For this variant, we replaced the ViT module with CNN with matching number of parameters. Tab.~\ref{tab:ablations} provides numerical comparisons of the two variants on image and 3D quality metrics. These quantitative and qualitative comparisons show that the ViT layers are important for creating more accurate 3D representations as well as achieving more accurate 2D image reconstruction. 

\paragraph{Effects of camera augmentation}
Fig.~\ref{fig:augmentation_ablation} compares the models trained with or without the camera augmentation for robustness to camera noise (also see the first row of Fig.~\ref{fig:comparison_1} for the results of the same subject without the camera noise). We fix the camera calibration and apply image space rotation, translation, and zoom to the input image, emulating the effect of inaccurate camera extrinsics and intrinsics. Although our model does not rely on any camera information for canonicalization, the result is not robust without the proposed camera augmentation. EG3D-PTI assumes a fixed image alignment used to train the GAN model and is very sensitive to small image misalignment in the input. Tab.~\ref{tab:ablations} provides numerical comparisons of our model with and without the proposed augmentation.

\revision{
\paragraph{Effects of fine-tuned synthetic discriminator.} We provide an additional ablation on the discriminator loss ($L_\text{adv}$ in Eqn.~\ref{eq:5}), which fine-tunes the pre-trained EG3D discriminator with EG3D-generated images. As seen in Tab.~\ref{tab:disc_ablation}, removing this discriminator loss results in a worse FID score. Moreover, as seen in Fig.~\ref{fig:disc_ablation}, the renderings of our proposed method are significantly sharper with the synthetic discriminator tuning. Please see Sec. A1 and Tab. A2 for attempts to train the discriminator with real images. 
}

\begin{figure}
    \centering
    \captionsetup[subfigure]{labelformat=empty} %
    \renewcommand\thefigure{\arabic{figure}} %
    \renewcommand{\thesubfigure}{}
    \begin{subfigure}{\linewidth}
        \includegraphics[width=\linewidth]{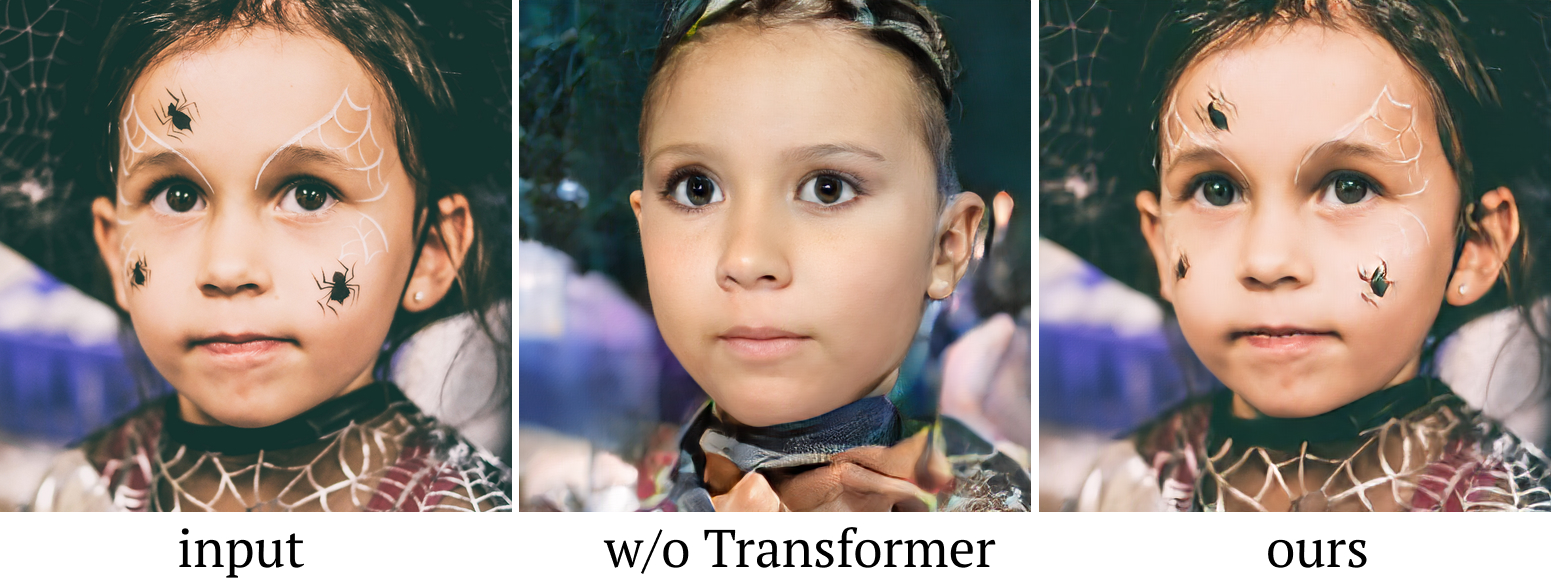}
        \caption{Fig. \thefigure.~Ablation study comparing our model with and without the proposed
Transformer modules. The model w/o Transformer replaces all Transformer
Blocks with resolution-preserving residual CNNs with similar parameters.
Credit to Kirill Chebotar.}
        \label{fig:transformer_ablation}
        \stepcounter{figure} %
    \end{subfigure}
    
    \vspace{1em} %
    \renewcommand{\thesubfigure}{}
    \begin{subfigure}{\linewidth}
        \includegraphics[width=\linewidth]{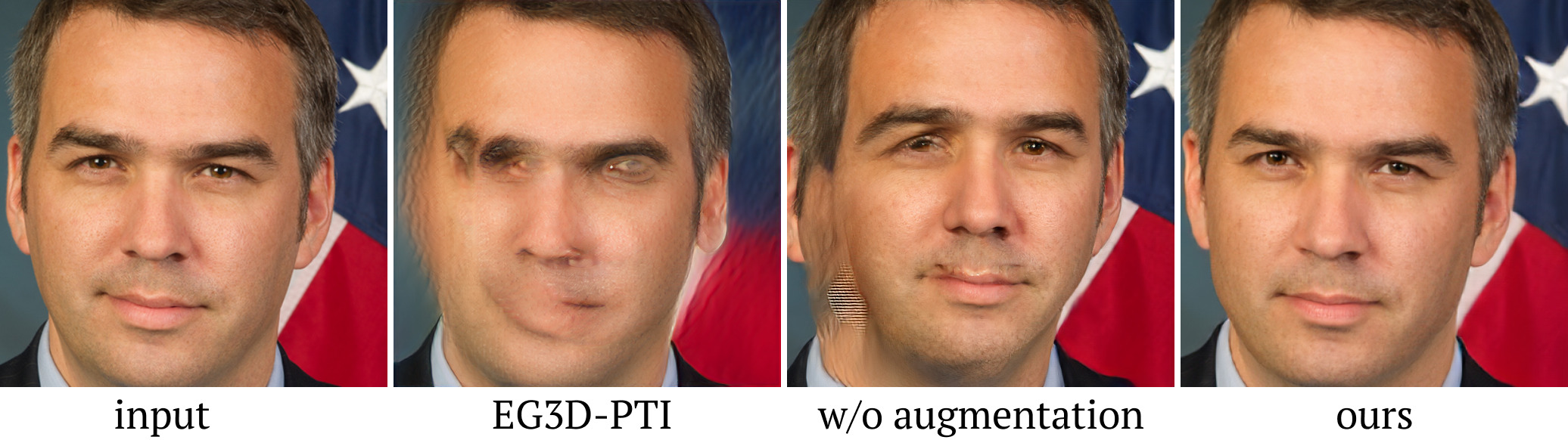}
        \caption{Fig. \thefigure.~Camera augmentation ablation study. Note that this is the same image as the first row of Fig.~\ref{fig:comparison_1} except rotated and cropped non-centrally. Without augmentation, our result exhibits artifacts when the input image has zoom or camera roll. Similarly, EG3D-PTI is also sensitive to the image misalignment, as the camera pose becomes noisy, while our method correctly canonicalizes the face. Images are cropped and aligned for visual consistency. Credit to U.S. Dept. of HUD.}
        \label{fig:augmentation_ablation}
        \stepcounter{figure} %
    \end{subfigure}
    
    \vspace{1em} %
    
    \renewcommand{\thesubfigure}{}
    \begin{subfigure}{\linewidth}
        \includegraphics[width=\linewidth]{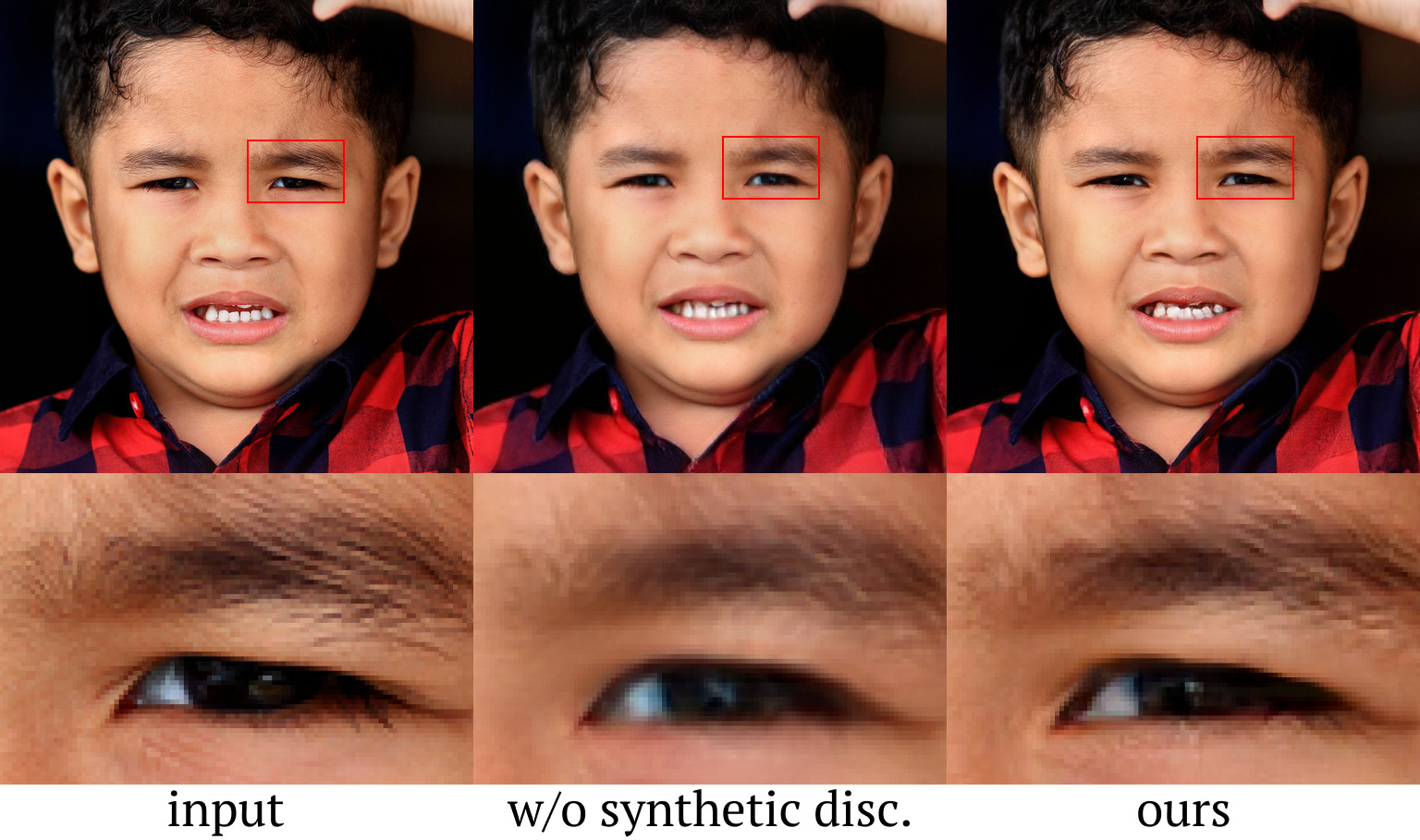}
        \caption{Fig. \thefigure.~Comparison between our model and an ablated model trained without the synthetic discriminator. Note the blurriness without the adversarial loss. Credit to Mohd Fazlin Mohd Effendy Ooi.}
        \label{fig:disc_ablation}
        \stepcounter{figure} %
    \end{subfigure}

    \addtocounter{figure}{-1}
    
    \captionsetup{labelformat=default}
    \renewcommand\thefigure{\arabic{figure}}
\end{figure}

\begin{table}[h]
\setlength{\tabcolsep}{4.5pt}
    \centering
		\small
	\caption{Ablation studies evaluating the proposed camera augmentation and the Transformer module. Without augmentation, the model acts as an autoencoder, mapping real images to arbitrary 3D representations that resemble the input (good ID score), but are not actually 3D (poor Pose score). Without a transformer, the encoder can canonicalize the inputs well (good Pose score), but cannot represent the details of the input (poor ID score). Our full method achieves both good Pose and ID scores with high reconstruction quality.  
    }
    \begin{tabular}{@{\hskip 1mm}l c c c c c c@{\hskip 1mm}}
				\toprule
					                & LPIPS $\!\downarrow$   & DISTS $\!\downarrow$         & Pose $\!\downarrow$   & ID $\!\uparrow$   & FID $\!\downarrow$ \\					
        \midrule
        No aug. 	                & 0.3846 		         & .1286                  	    & 0.1758                & \textbf{0.5359}   & \textbf{3.42} 	             \\
        No Transformer 	            & 0.5419 		         & .1650                   		& \textbf{0.0426}       & 0.1906            & 11.5 	             \\
        Ours  		      	& \textbf{0.2894}        & \textbf{.1053}          	    & 0.0461                & 0.5230            & 4.45        \\
				\bottomrule
    \end{tabular}
    \quad
    \label{tab:ablations}
\setlength{\tabcolsep}{6pt}

\end{table}

\begin{table}[h]
    \centering
  \small
    \caption{Comparison in FID between our model and an ablated model without the synthetic discriminator.}
    \begin{tabular}{@{\hskip 1mm}l c@{\hskip 1mm}}
				\toprule
		FID $\!\downarrow$ 		& FFHQ   \\
        \midrule
        w/o synthetic disc. 	    & 7.71 					  \\
        Ours  		      	    & \textbf{4.45}   \\
				\bottomrule
    \end{tabular}

    \label{tab:disc_ablation}
\end{table}

\subsection{Application: real-time 3D telepresence}
\begin{figure}
    \centering
    \includegraphics[width=0.95\linewidth]{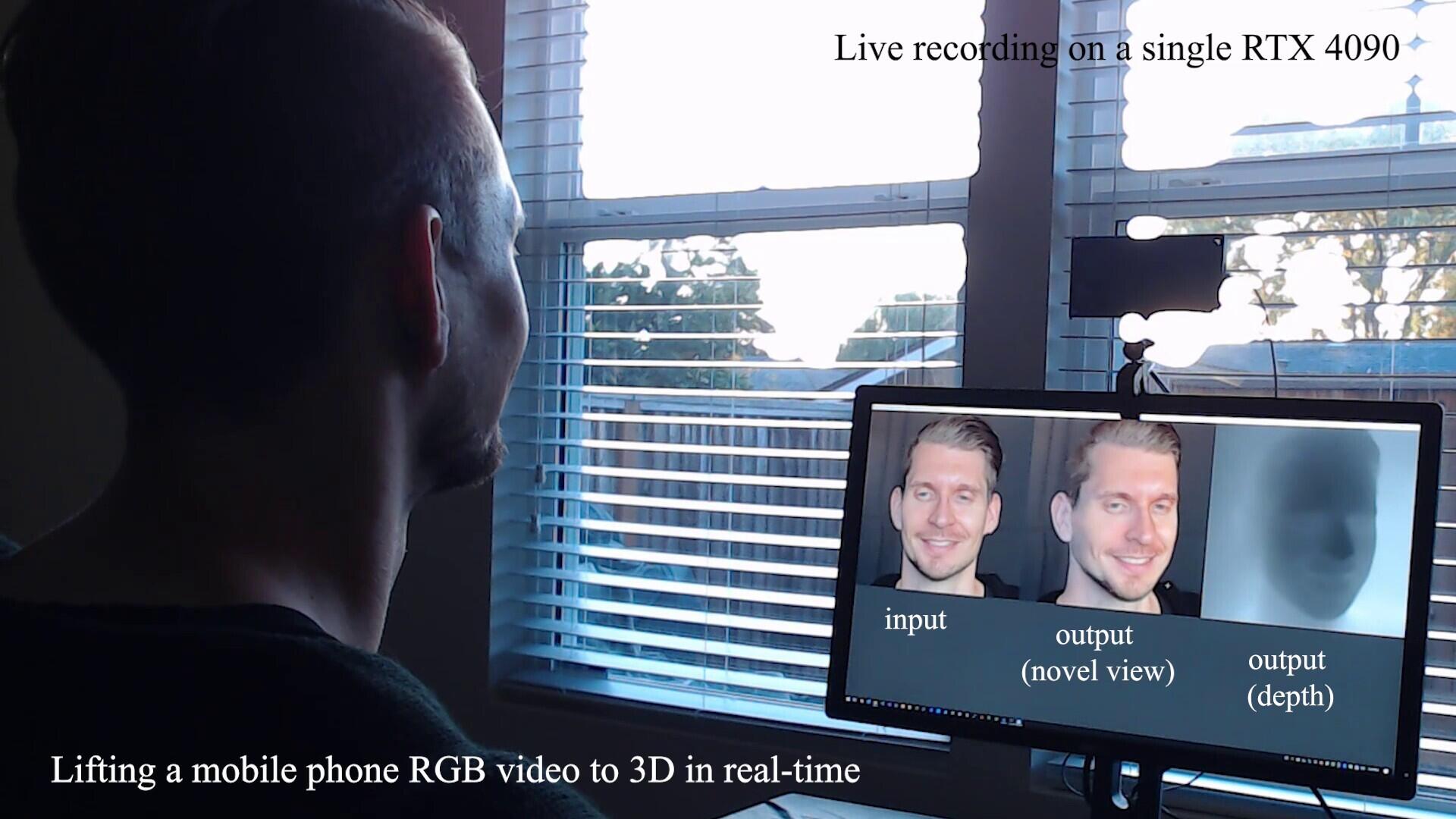}
    \caption{Our system applied to create a 3D telepresence live from a monocular RGB input. Please see the supplement video for the live demonstration.}
    \label{fig:3Dvidconf}
\end{figure}

We apply our method for lifting a monocular RGB video input to 3D in real-time, as would be needed for 3D telepresence. Our method processes the video frame by frame. Despite being trained on individual frames of synthetic data and processing the input video in a frame to frame fashion, our method can provide reasonable temporal consistency. Please refer to the teaser Fig.~\ref{fig:teaser} (bottom right) for the output from our lightweight model as well as video examples from the supplement.
Fig.~\ref{fig:3Dvidconf} shows our system set up and running off of a desktop with a single RTX 4090. 
Our method can lift a monocular RGB video frame from a mobile phone to 3D in real-time.

\section{Discussion}
\label{sec:discussion}
\revision{
\paragraph{Limitation.}
When the input is a strong profile view (e.g., 60 degrees yaw angle), our method may struggle with properly canonicalizing the input, as it is highly out-of-distribution with respect to EG3D-generated images and FFHQ. Please see Figs.~\ref{fig:pitch} and \ref{fig:yaw} for various challenging levels of pitch and yaw for input images. While our method can predict a canonicalized 3D representation without requiring camera poses as input, the rendered image may be slightly misaligned when compared to the input view (see Fig.~A8 in the supplement for the detailed analysis) possibly due to the combination of the imperfect canonicalization and noisy camera poses from an off-the-shelf pose estimator. Finally, although our method can provide reasonable temporal consistency when applied to a video in a frame-by-frame fashion, temporal inconsistencies remain as the canonicalizations change slightly per frame, and the predicted camera poses are entirely independent.  
}

\paragraph{Future work.}
In the future, combining our method with camera pose optimization ~\cite{ko20233d3dganinversion} may lead to more accurate 3D reconstruction and camera pose estimation. Additionally, jointly predicting the camera poses and triplanes in an autoregressive or recurrent context \cite{srivastava2015unsupervised, kalchbrenner2017video, shi2015convolutional} may result in more consistent frame-by-frame results. Next, it would be interesting to incorporate real images in the training as our preliminary attempts did not yield improvements. Finally, as our pipeline does not necessarily assume any category-specific priors, we can view it as a general method to distill the knowledge of a 3D GAN into a feedforward encoder. Thus, extending 3D GANs to more general scenes~\cite{imagenet3d} may allow our pipeline to create 3D representations of arbitrary scenes in the future. Specifically extending our work to handle hands or the full body, is of interest for real-time telepresence applications.

\paragraph{Conclusion.}
We proposed a one-shot encoder-based framework to lift a single RGB image to 3D in real-time and demonstrated our method, trained entirely from synthetic data, can handle challenging (even out-of-domain) real-world images.
We believe that this opens up possibilities for accessible 3D reconstructions of real-world objects and interactive 3D visualization from a picture.  %

\section*{Acknowledgements}
We thank David Luebke, Jan Kautz, Peter Shirley, Alex Evans, Towaki Takikawa, Ekta Prashnani and Aaron Lefohn for feedback on drafts and early discussions. We acknowledge the significant efforts and suggestions of the reviewers. For allowing the use of video, we thank Elys Muda. This work was funded in part at UCSD by ONR grants N000142012529, N000142312526, an NSF
graduate Fellowship, a Jacobs Fellowship, and the Ronald L. Graham
chair. Manmohan Chandraker acknowledges support of NSF IIS 2110409. Koki Nagano and Eric Chan were partially supported by DARPA’s Semantic Forensics (SemaFor) contract (HR0011-20-3-0005). The views and conclusions contained in this document are those of the authors and should not be interpreted as representing the official policies, either expressed or implied, of the U.S. Government. Distribution Statement ``A'' (Approved for Public Release, Distribution Unlimited).

\bibliographystyle{ACM-Reference-Format}
\bibliography{egbib}

\end{document}

\title{Supplementary Material \\Real-Time Radiance Fields for Single-Image Portrait View Synthesis}
\makeatletter
    \renewcommand\AB@affilsepx{ \hphantom{---} \protect\Affilfont}
\makeatother

\newcommand*\samethanks[1][\value{footnote}]{\footnotemark[#1]}

\author[2]{Alex Trevithick\thanks{This project was initiated and substantially carried out during an internship at NVIDIA.}}
\author[1]{Matthew Chan}
\author[1]{Michael Stengel}
\author[3]{Eric Ryan Chan\samethanks[1]}
\author[1]{Chao Liu}
\author[1]{Zhiding Yu}
\author[1]{\\Sameh Khamis}
\author[2]{Manmohan Chandraker}
\author[2]{Ravi Ramamoorthi}
\author[1]{Koki Nagano}

\affil[1]{NVIDIA}
\affil[2]{University of California, San Diego}
\affil[3]{Stanford University}

\maketitle

In this supplement, we first provide the additional results including additional evaluations and comparisons in Sec.~\ref{sec:results}. We provide the implementation details of our models, including architecture details, camera augmentation, training details, and hyper parameters in Sec.~\ref{sec:implementation}. We also provide further experiment details in Sec.~\ref{sec:experiment_details}. Finally, we discuss the limitations of our work in Sec.~\ref{sec:discussion}. We encourage the readers to view our accompanying videos in the supplement, which include the additional visual comparisons, results, and live demonstration of the novel view synthesis from a video input.  

\section{Additional results}
\label{sec:results}

\subsection{Additional qualitative results}
\begin{figure*}
    \centering
    \includegraphics[width=0.93\textwidth]{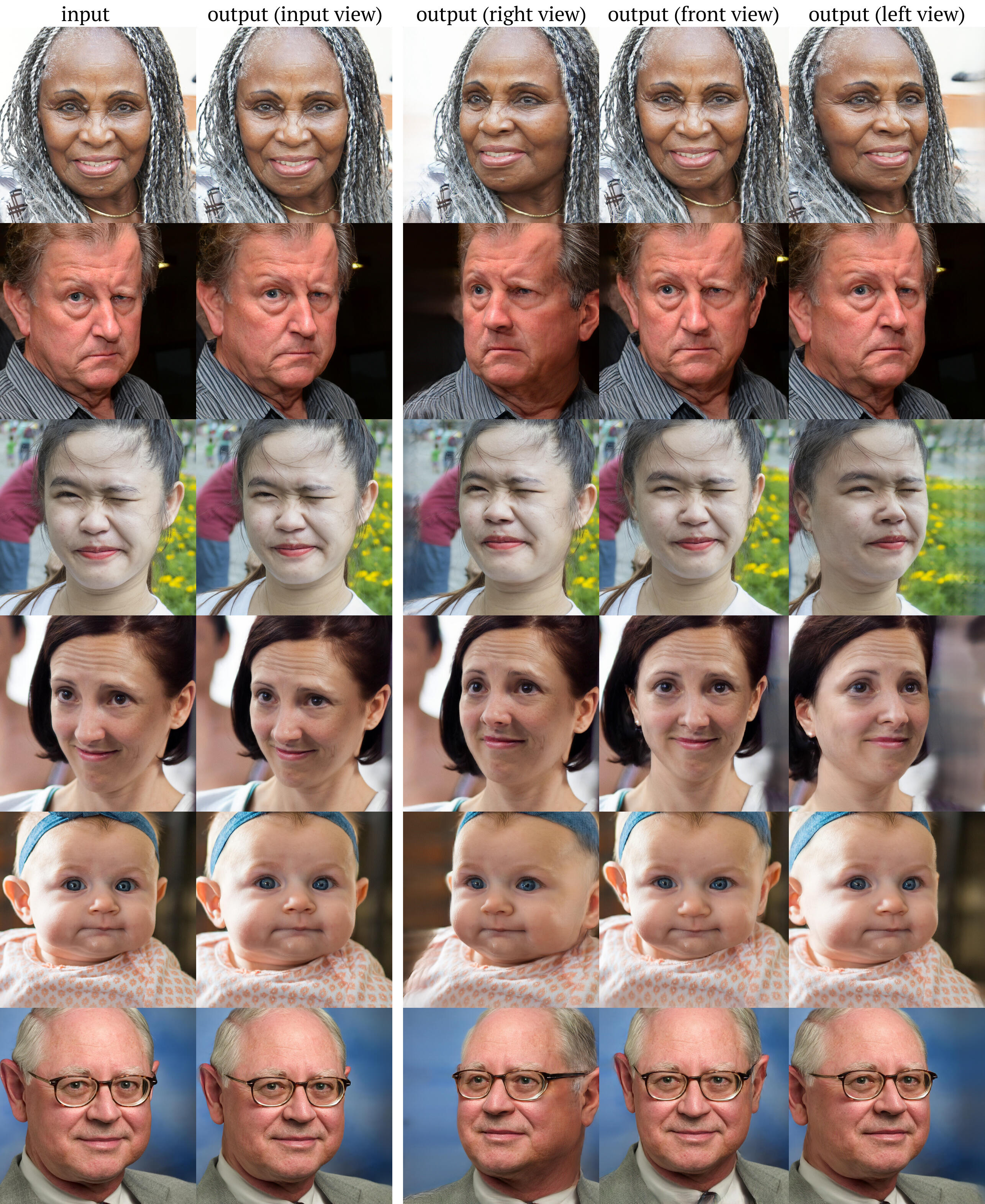}
    \caption{Additional qualitative results generated by our method on FFHQ. Credits to USAID | Southern Africa, TimothyJ, toan đào song, Travis Rock, Curt Mills, UGA CAES/Extension.}
    \label{fig:supp_qualitative_FFHQ}
\end{figure*}

\begin{figure*}
    \centering
    \includegraphics[width=0.95\textwidth]{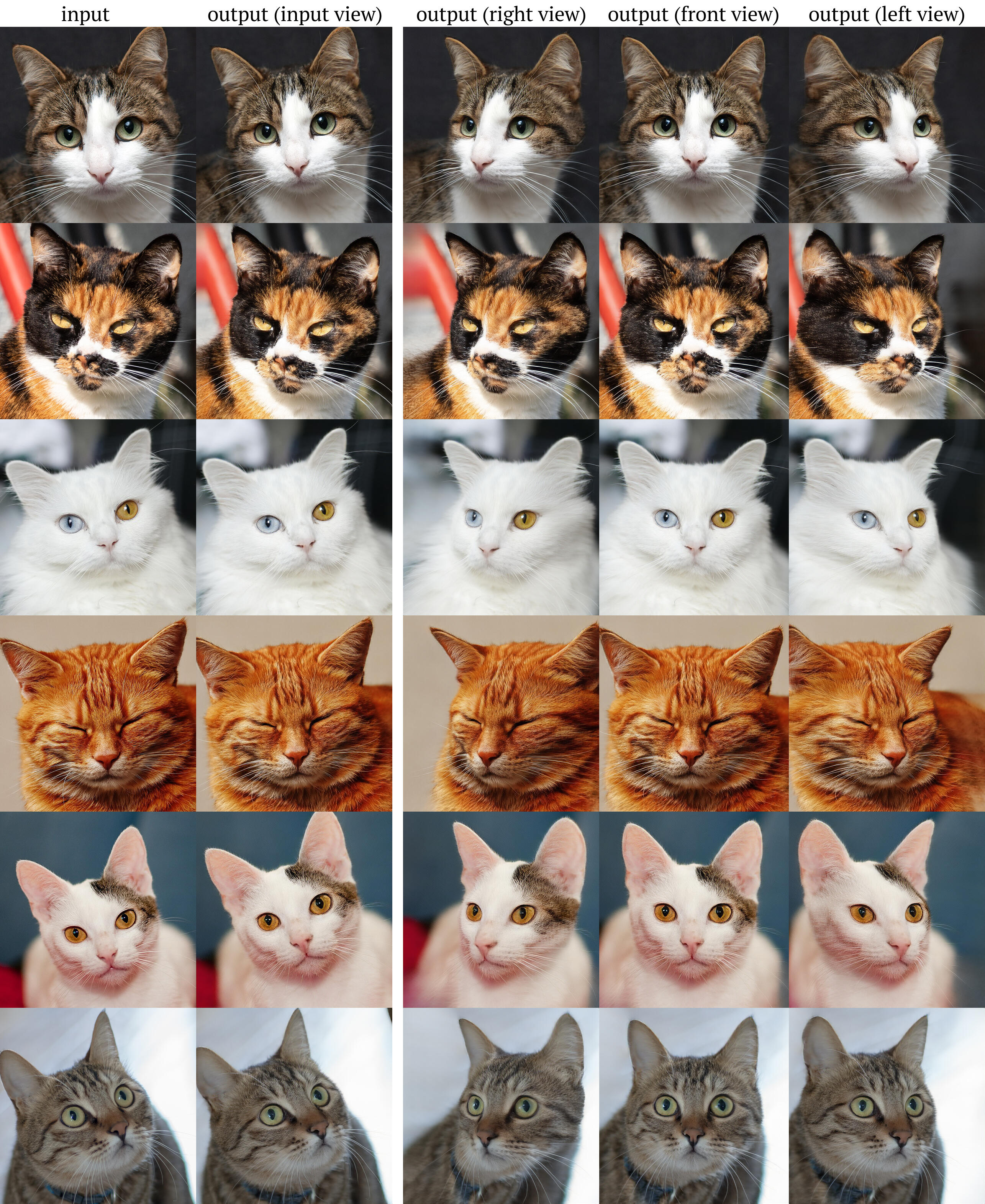}
    \caption{Additional qualitative results generated by our method on AFHQ. }
    \label{fig:supp_qualitative_AFHQ}
\end{figure*}

We provide additional qualitative results generated from a single input image from FFHQ in Fig.~\ref{fig:supp_qualitative_FFHQ} and AFHQ in Fig.~\ref{fig:supp_qualitative_AFHQ}. %
Fig.~\ref{fig:supp_qualitative_FFHQ} shows that our method can handle complex hairstyles (first row), and asymmetric facial expressions (second and third rows). Fig.~\ref{fig:supp_qualitative_AFHQ} shows our method can handle unconstrained poses of cats present in the portraits as well as a wide variety of their textures. 

\subsection{Qualitative comparisons to ~\cite{ko20233d3dganinversion}}
\begin{figure}
    \centering
    \includegraphics[width=\linewidth]{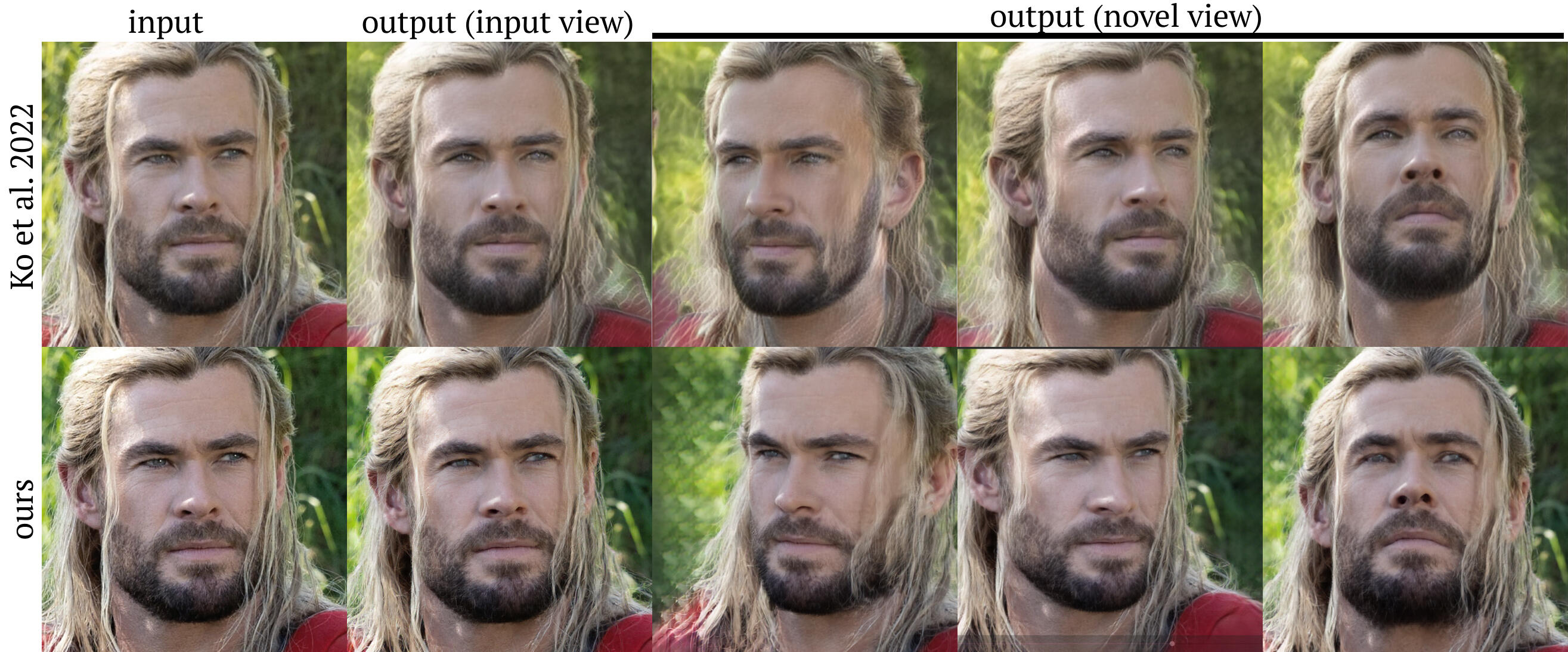}
    \caption{Qualitative comparisons to the concurrent work\cite{ko20233d3dganinversion} that relies on test-time camera optimization and generator weights tuning.}
    \label{fig:comparison_ko}
\end{figure}

In Fig.~\ref{fig:comparison_ko}, we provide comparisons to the state-of-the-art 3D GAN inversion work by ~\cite{ko20233d3dganinversion}.
While their method needs test-time optimization for the camera parameters and generator tuning, our method can process an unposed input in one-shot. 

\begin{figure}
    \centering
    \includegraphics[width=\linewidth]{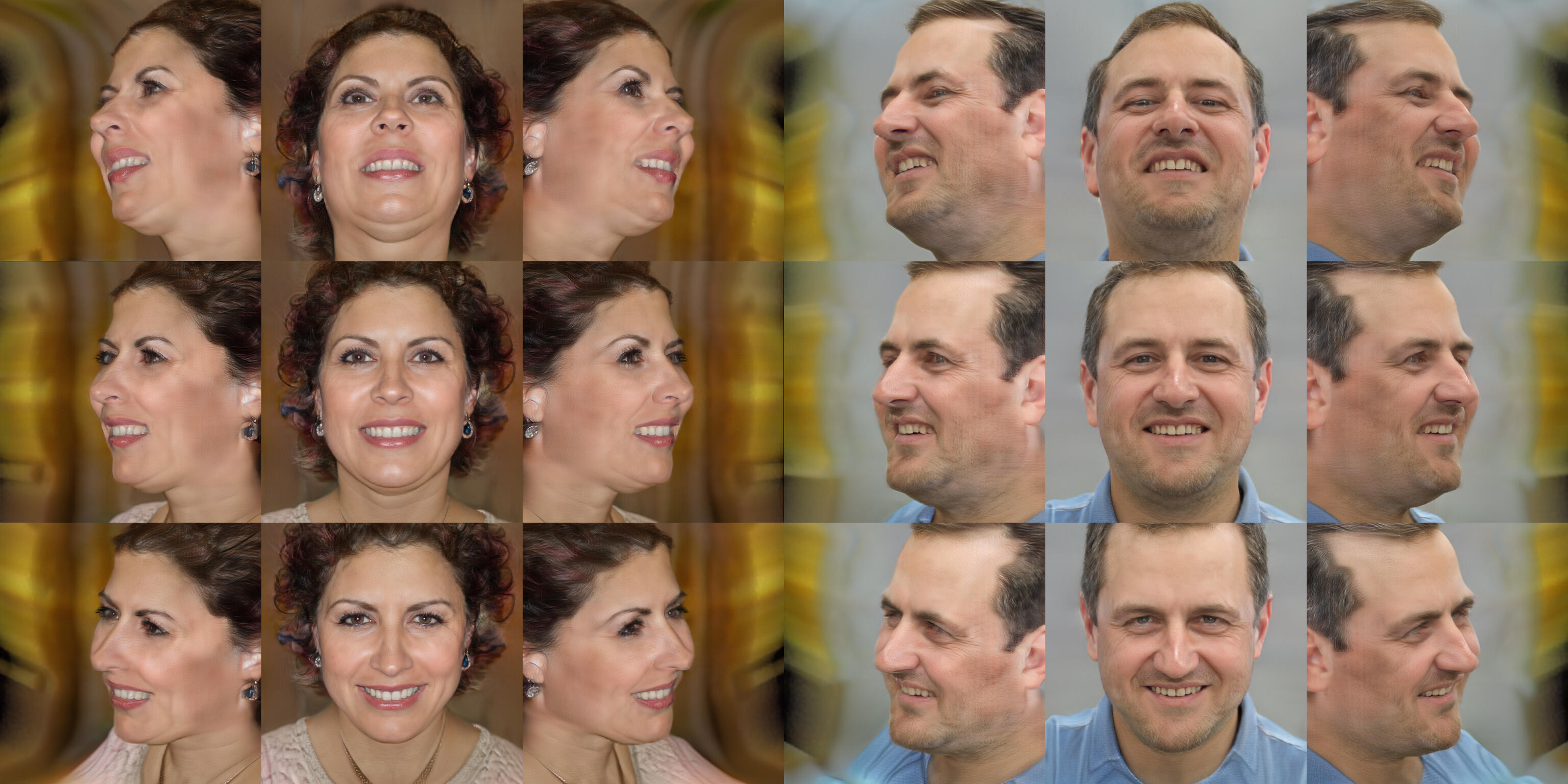}
    \caption{Visualization of the limits in pitch and yaw of the camera pose distribution for synthetic input images used to supervise our model.}
    \label{fig:synthetic_pose}
\end{figure}

\begin{figure}
    \centering
    \includegraphics[width=\linewidth]{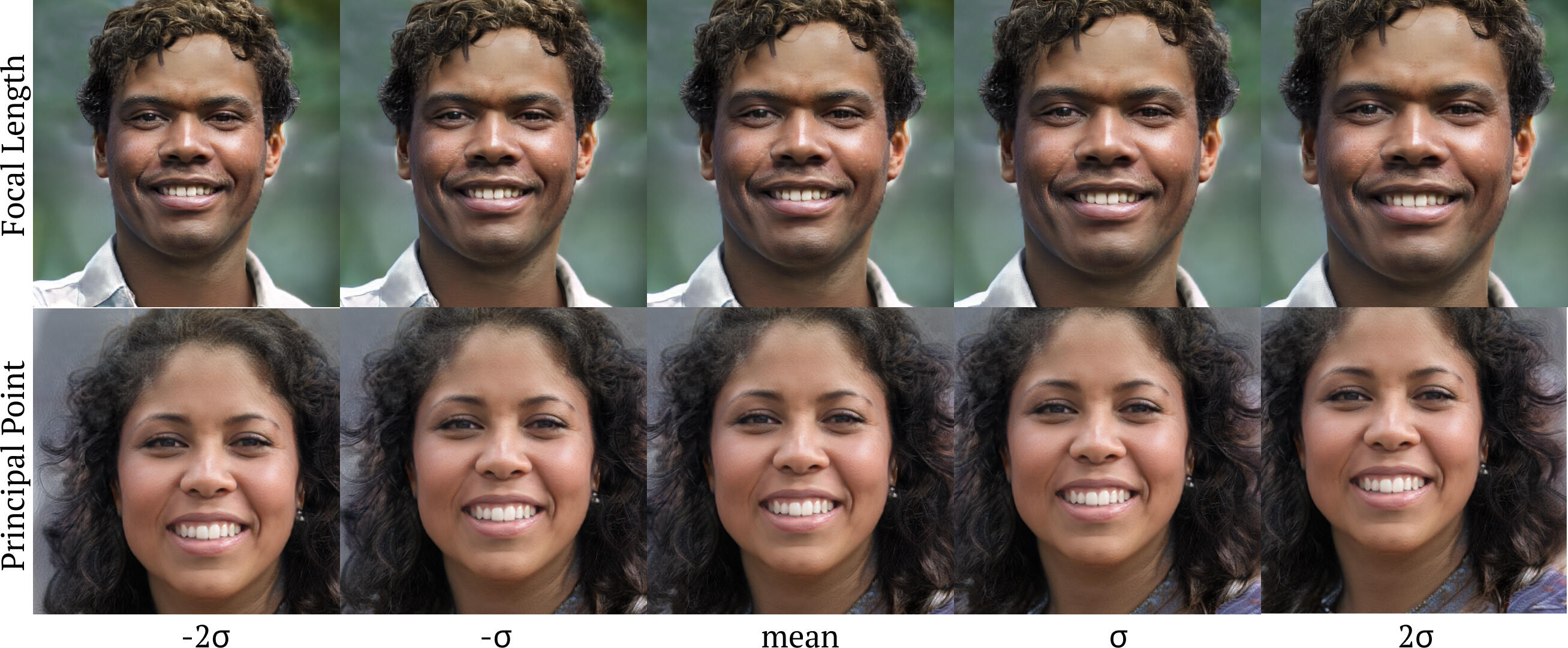}
    \caption{Visualization of two sigmas of noise in the principal point and focal length used for camera augmentation during our training.}
    \label{fig:synthetic_focal}
\end{figure}

\subsection{Additional comparisons}
\begin{figure*}
    \centering
    \includegraphics[width=\textwidth]{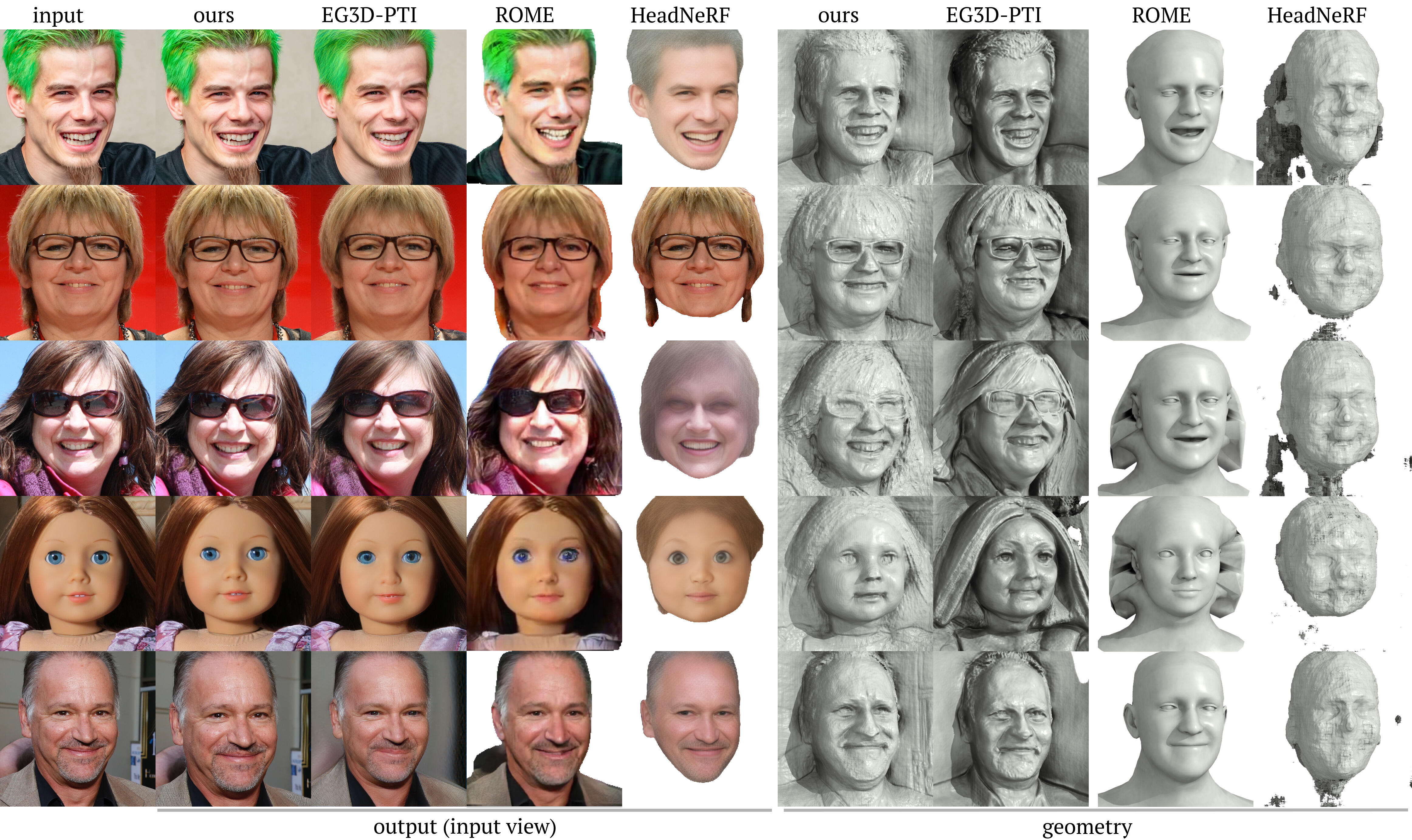}
    \caption{Additional qualitative comparisons against baselines on input view reconstruction and geometry. Credit to Steffen Geyer, Force Ouvrière, Matt Hamm, scarlett1854, The Society of Motion Picture and Television Engineers.}
    \label{fig:sup_ffhq_comparison}
\end{figure*}

We provide additional comparisons to HeadNeRF, ROME, and EG3D-PTI in Fig.~\ref{fig:sup_ffhq_comparison}.
 HeadNeRF only reconstructs the head region and struggles to reconstruct out-of-domain hair color (first row). ROME reconstructs the foreground image well, but requires background segmentation and the geometry does not fully capture the hairstyles and eyeglasses (second and third rows). EG3D-PTI reconstructs full RGB images and geometry, but occasionally produces distorted 3D shapes (first row, better viewed in 3D in the accompanying video) when the input view is non-frontal. Our method produces consistent image and geometry reconstruction quality across the variety of inputs including a non-realistic human image (fourth row).

\subsection{Percentile results based on LPIPS}
\begin{figure*}
    \centering
    \includegraphics[width=0.82\linewidth]{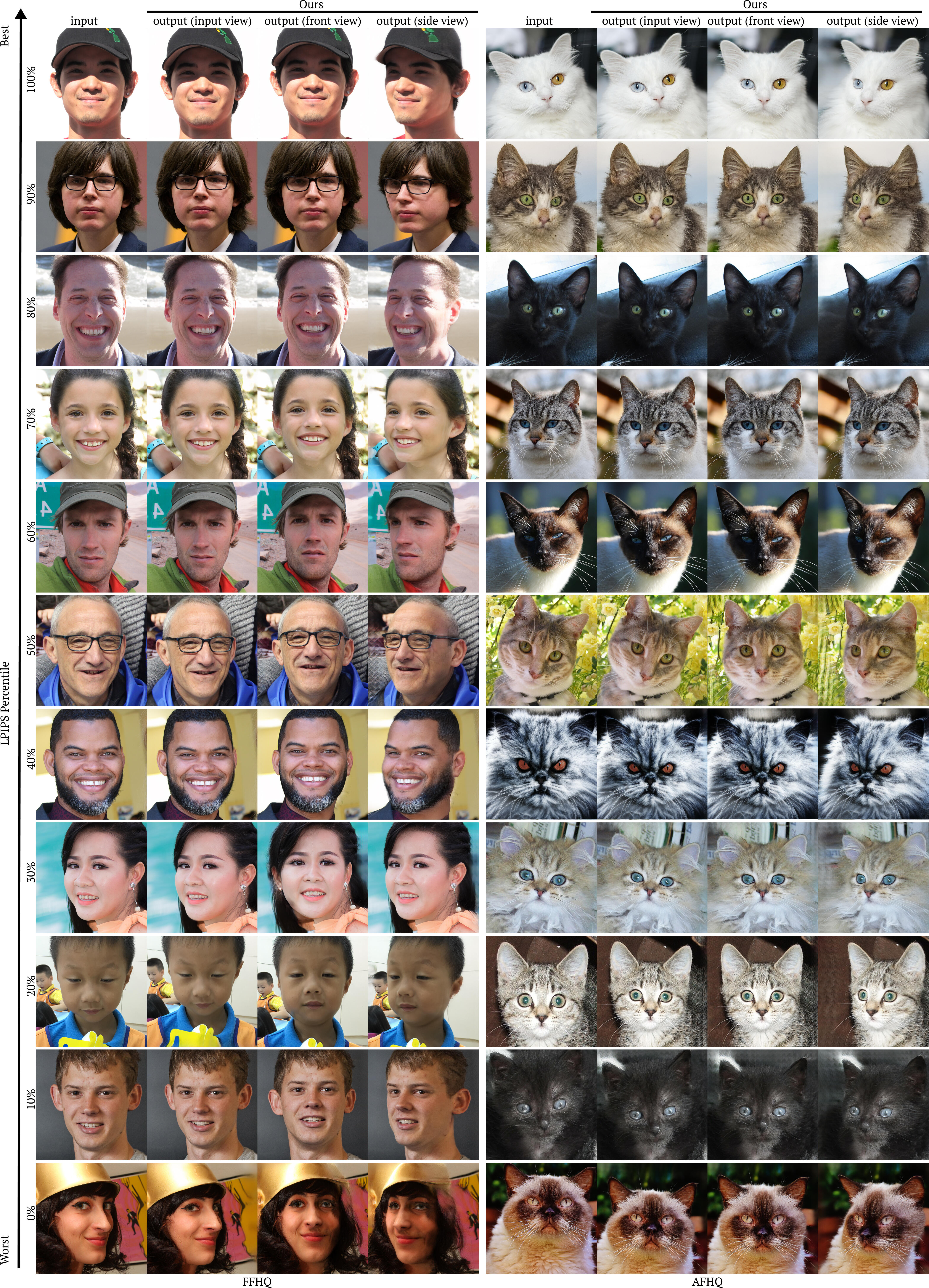}
    \caption{Results generated by our method on FFHQ and AFHQ shown in the order of percentile. Note that percentiles for FFHQ are calculated with alignment, whereas AFHQ percentiles are calculated without alignment from a test set. Credits to yasminehabib, Rutgers Council on Public and Internation Affairs, davitydave, Laity Lodge Family Camp, Houston Marsh, Ordiziako Jakintza Ikastola, Edgar Caraballo, NGÁO STUDIO, Debbie, WorldSkills UK, Craig Duffy.}
    \label{fig:percentile_LPIPS}
\end{figure*}

In Fig.~\ref{fig:percentile_LPIPS}, we show our results on FFHQ and AFHQ shown in the order of the LPIPS percentile scores. 
For FFHQ, we use the same randomly selected 500 FFHQ test set described in the main manuscript and for AFHQ, we randomly selected 485 images for which we computed the LPIPS scores. The percentile results preferred by the LPIPS scores show that our method can demonstrate consistent quality for the large portion of the test images. 

\begin{figure}
    \centering
    \includegraphics[width=\linewidth]{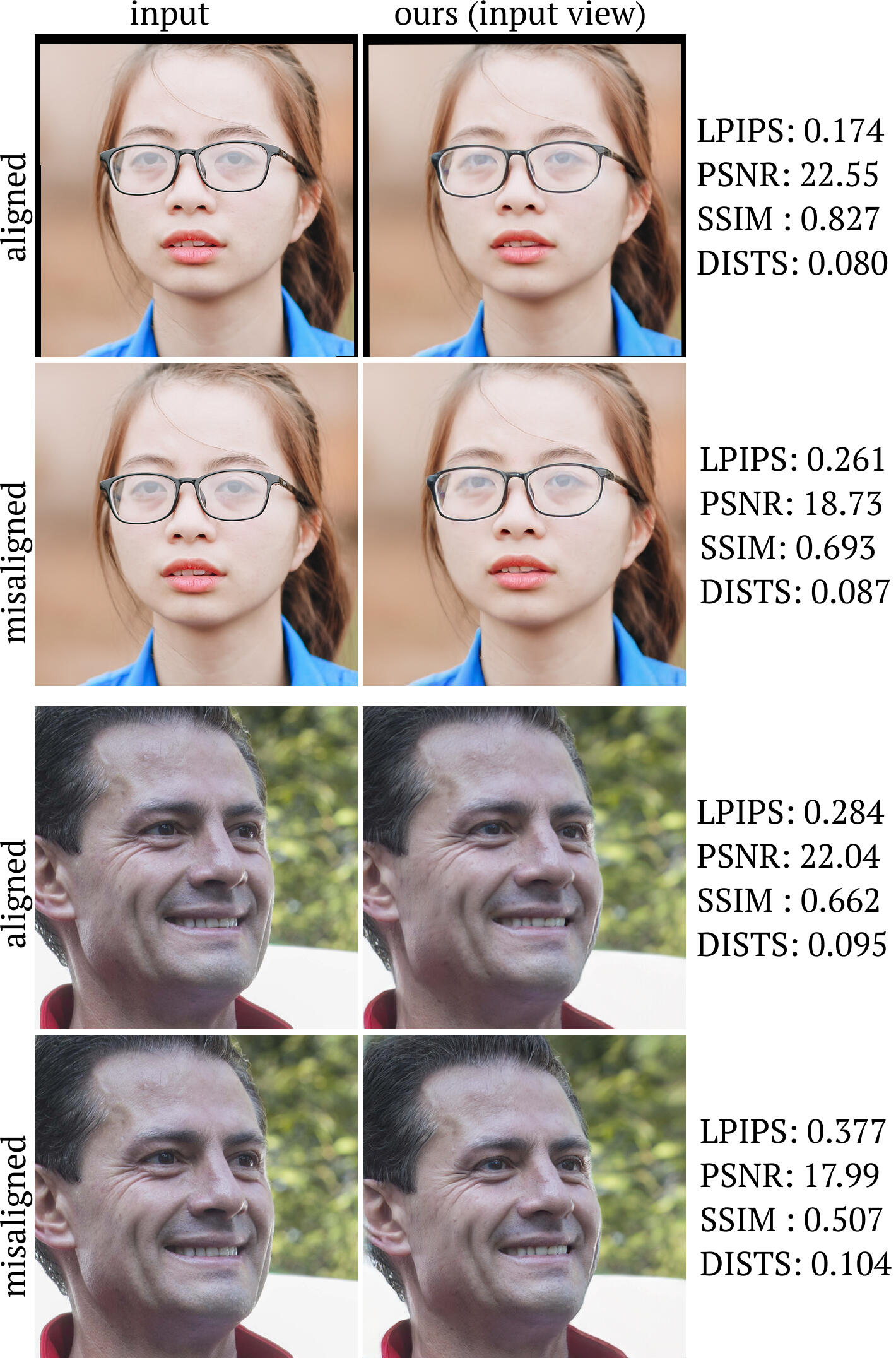}
    \caption{LPIPS, DISTS, PSNR, and SSIM scores computed on images that have a small misalignment. Previous work~\cite{ding2020iqa} reported that LPIPS can tolerate small geometric misalignment better than PSNR and SSIM; DISTS is most robust to small image misalignment. Credits to Dong Quang, Presidencia de la Republica Mexicana.}
    \label{fig:misalignment}
\end{figure}

\paragraph{PSNR and SSIM on misaligned images.}
We provide our analysis on PSNR and SSIM metrics on images when images are aligned and when images have a small misalignment in Fig.~\ref{fig:misalignment}. 
While LPIPS scores can tolerate a small image misalignment (little change when images are aligned or misaligned), the PSNR and SSIM scores significantly change, which make these metrics unreliable for our tasks when the reconstructed images are not perfectly pixel-to-pixel aligned. The issues of PSNR and SSIM scores sensitivity under geometry transformation are reported by previous work~\cite{ding2020iqa}. The DISTS~\cite{ding2020iqa} metric can also tolerate slight misalignment.

\begin{table}[t]
    \centering
  \small
    \caption{Comparisons to an \textit{unconditional} reference on FID evaluated over 50K images of FFHQ and 10k images of AFHQ (including horizontal flips). $^\dag$Using transfer learning from a pretrained FFHQ model.  %
    }
    \begin{tabular}{@{\hskip 1mm}l c c@{\hskip 1mm}}
				\toprule
		FID $\!\downarrow$ 		& FFHQ              & AFHQ\\
        \midrule
        EG3D 	    & 4.05 				& 2.88$^\dag$ 	            \\
        Ours  		      	    & \textbf{3.48}     & 2.39$^\dag$  \\
        Ours (LT)  		        & 4.25              & \textbf{2.11}$^\dag$ \\
				\bottomrule
    \end{tabular}

    \label{tab:FID_comparisons}
\end{table}

\subsection{Evaluation of FID}
Tab.~\ref{tab:FID_comparisons} provides comparisons on FID calculated over 50K images from FFHQ and 10K images from AFHQ. Our lightweight model "Ours (LT)" produces competitive FID scores to our full model ("Ours"). 

\subsection{Ablation study}
We provide additional ablation studies concerning the importance of the training dataset size and describe our preliminary attempt to incorporate real images into the training. 

\begin{table}[t]
\setlength{\tabcolsep}{4.5pt}
    \centering
		\small
    \caption{Additional ablation study comparing variants of our models. "10K" refers to when we pre-compute 10K triplanes (subjects) and generate supervising views on the fly using EG3D.  
    "w/real data" shows preliminary results of our initial attempt to incorporate real images in the training. 
    }  
    \begin{tabular}{@{\hskip 1mm}l c c c c c c@{\hskip 1mm}}
				\toprule
					                & LPIPS $\!\downarrow$   & DISTS $\!\downarrow$         & Pose $\!\downarrow$   & ID $\!\uparrow$   & FID $\!\downarrow$ \\					
        \midrule
        10K 	                & \textbf{0.2797} 		         &\textbf{0.995}                  	    & \textbf{0.0458}                & 0.5000   & 4.60 	             \\
        w/real data 	                & 0.3060 		         & 0.1125                  	    & 0.0539                & 0.4556   & 6.15	             \\
        Ours  		      	& 0.2894        & .1053          	    & 0.0461                & \textbf{0.5230}            & \textbf{4.45}        \\
				\bottomrule
    \end{tabular}
    \quad

    \label{tab:sup_ablation}
\setlength{\tabcolsep}{6pt}
\end{table}

\begin{figure}
    \centering
    \includegraphics[width=\linewidth]{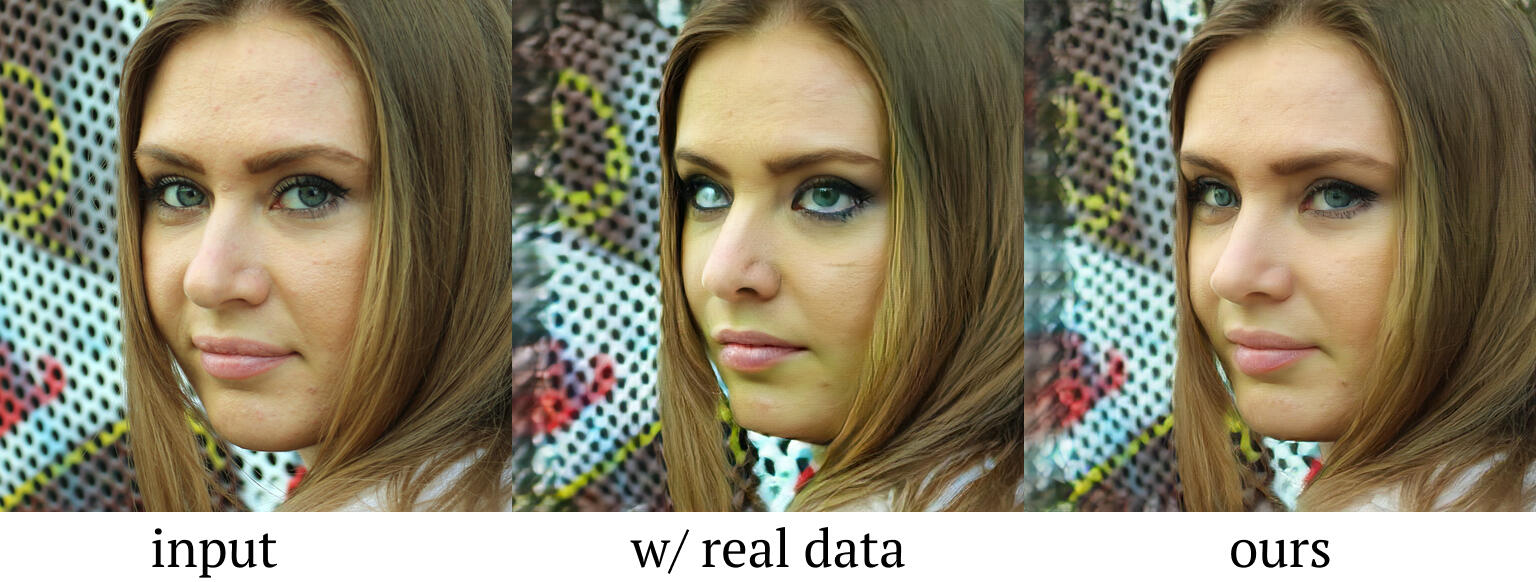}
    \caption{Comparisons showing the output of our initial attempt in incorporating real images. Credit to Mario Krajčír.}
    \label{fig:ablation_real_data}
\end{figure}

\paragraph{Adding real data to the training.}
We attempted to incorporate real data into the training pipeline in a variety of ways, but each one proved unsuccessful. Our most succesful attempt was to train the real part of the discriminator with images from FFHQ (using the same conditioning as the original EG3D) and add significant noise to the discriminator pose conditioning. Tab.~\ref{tab:sup_ablation} shows the results of our preliminary attempt to incorporate real images in the training. As can be seen in Fig. \ref{fig:ablation_real_data}, even this method fails to reconstruct the input image faithfully. 
\paragraph{Size of training data.}
We additionally performed an ablation on the number of subjects in the training set. To do so, we chose 10k latent codes from EG3D and rendered images from only these. We found that training with 10k subjects with on the fly supervising view generation (theoretically each subject has infinite views to supervise) performs similarly to our method which synthesizes new identities on the fly, as seen in Tab.~\ref{tab:sup_ablation}. We hypothesize that this is because the number of subjects is similar to the datasets, such as VGGFace2 \cite{cao2018vggface2} (9K subjects) and CASIA-WebFace~\cite{Yi2014LearningFR} (10K subjects), used to train a one-shot face recognition model.%

\subsection{Additional Applications}

\paragraph{Portrait frontalization.}
Our method can be applied to portrait face frontalization, which is useful for 3D reconstruction and avatar digitization~\cite{nagano2019deepfacenorm}. Please see the examples in Figs.~\ref{fig:supp_qualitative_FFHQ} (4th column),~\ref{fig:supp_qualitative_AFHQ} (4th column),~\ref{fig:percentile_LPIPS} (3rd and 7th columns).%

\section{Implementation details}
\label{sec:implementation}
We implement our framework in PyTorch on top of the official EG3D codebase (\url{https://github.com/NVlabs/eg3d}). 
\paragraph{EG3D pre-trained model.}
For human faces, we use the EG3D model trained on the FFHQ dataset (\textit{ffhqrebalanced512-128.pkl}).
To simplify our encoder training supervision, we replaced the latent code $W$ injected to the StyleGAN2-based super-resolution layer with a constant 1 and fine tuned the entire EG3D models on FFHQ for additional 6.8 M images of training. This resulted in the FID score 4.05 for FFHQ as reported in the main manuscript. For cats faces, we performed transfer learning~\cite{Karras2020ada} from this FFHQ checkpoint and trained additional 3.2M images on the cat split of AFHQv2, which resulted in the FID score 2.88, again reported in the main manuscript. Please refer to the samples of synthetic data generated by the EG3D model in Figs.~\ref{fig:synthetic_pose} and \ref{fig:synthetic_focal}.

\begin{figure*}[hbt!]
  \includegraphics[width=\linewidth]{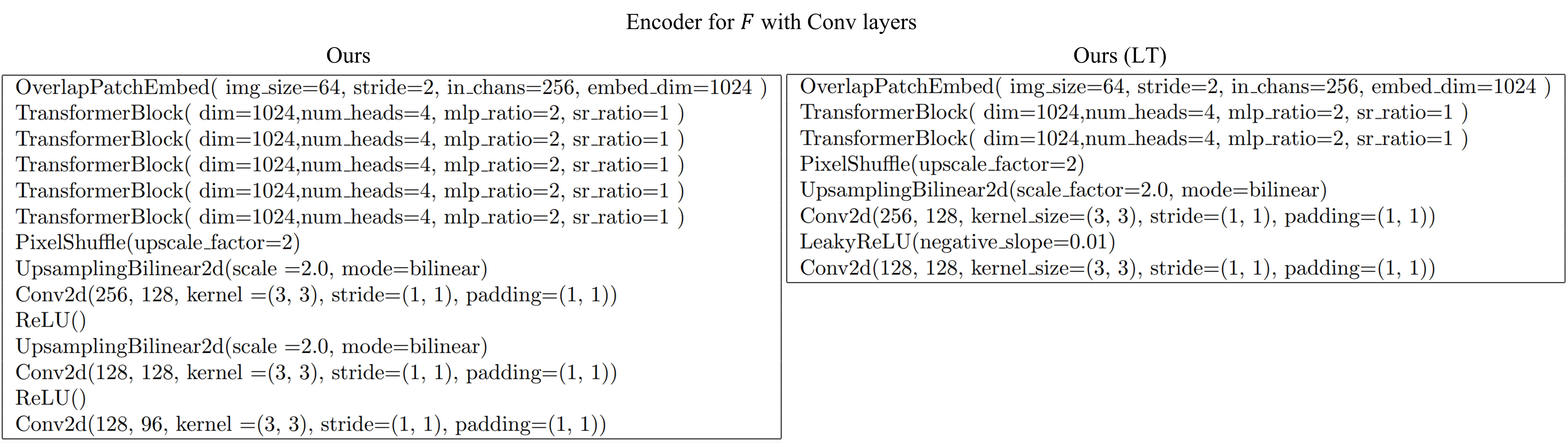}
  \caption{Details of the hybrid convolutional-transformer architecture which decodes the DeepLabV3 features before being concatenated with the high-resolution image features later on.}
   \label{fig:E_low}

\end{figure*}

\begin{figure*}[hbt!]
  \includegraphics[width=\linewidth]{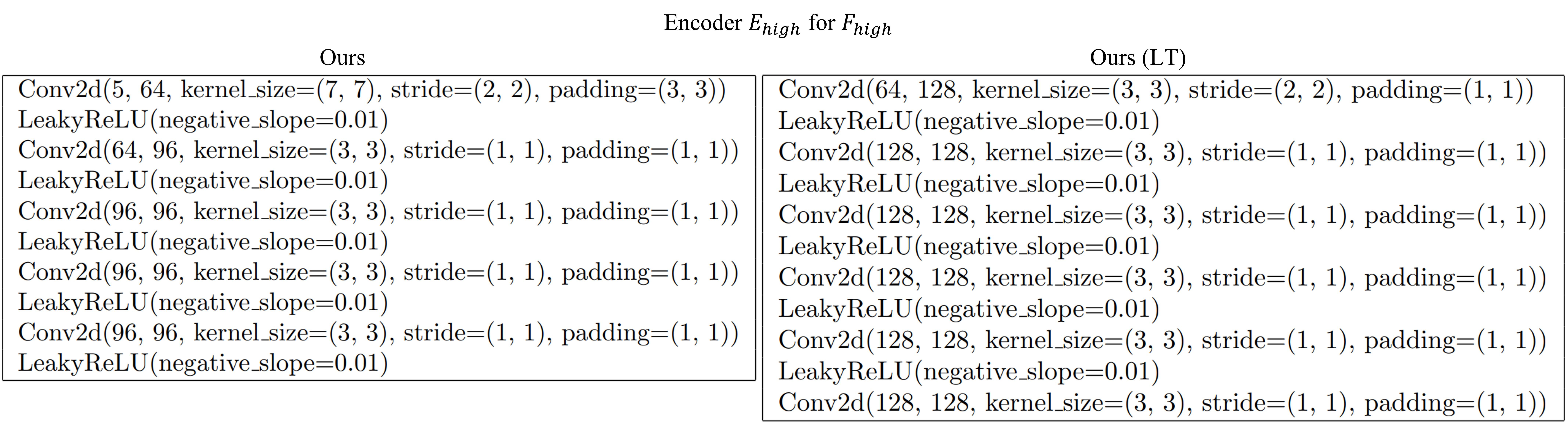}
  \caption{Details of $E_\text{high}$ which maps the input image to a high-resolution feature map.}
  \label{fig:E_high}
\end{figure*}

\begin{figure*}[hbt!]
  \includegraphics[width=\linewidth]{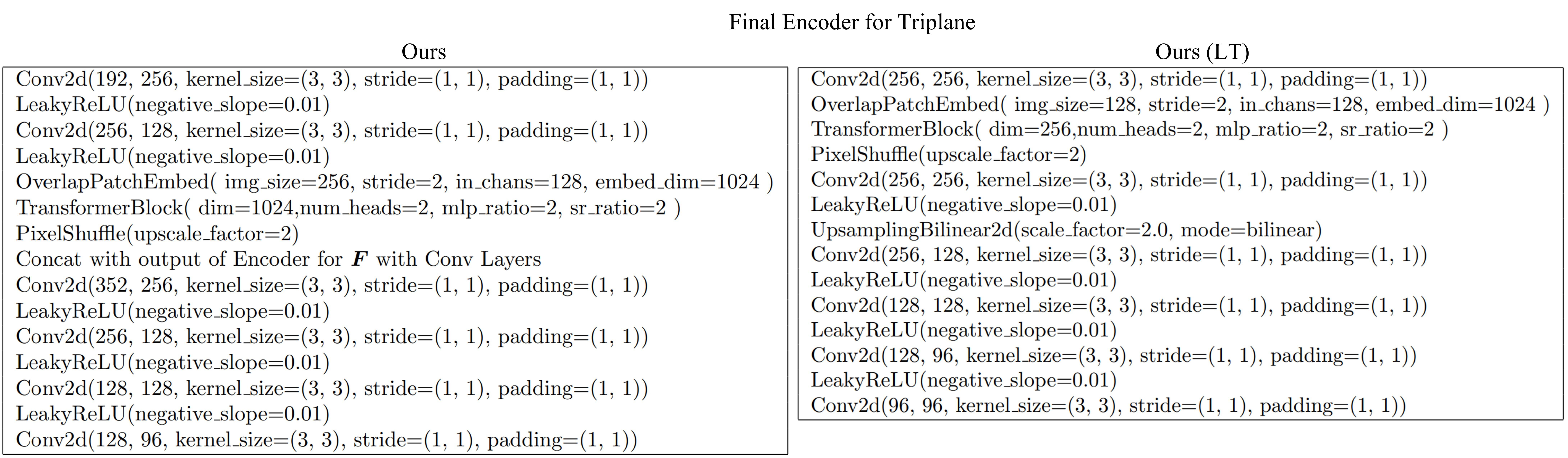}
  \caption{Details of the hybrid convolutional-transformer architecture which decodes the concatenated transformer features and high-resolution image features directly into a triplane representation.}
\label{fig:processor}
\end{figure*}

\paragraph{Encoder for $\bm F_\text{low}$.}
We modify the first layer of the DeepLabV3 \cite{chen2017rethinking} architecture from the Pytorch Segmentation Models repo \cite{Iakubovskii:2019} by concatenating the 2D pixel coordinate of each pixel, so that the input is 5 channels. We also remove all instances of batch norm (reintroducing the biases in all of the convolutional layers). Otherwise, we use the standard encoder-decoder as implemented with a ResNet34 encoder. We take the feature map output of the decoder of DeepLabV3 (the layer before bilinear upsampling and segmentation head). As seen in the top half of the pipeline figure in the main paper, this gives us a feature map $\bm F_\text{low}$.
\paragraph{Encoder for $\bm F$ with Conv Layers.}
$\bm F_\text{low}$ is fed to a hybrid convolutional-transformer architecture. We will denote OverLapPatchEmbed as the patchwise embedding from Segformer \cite{xie2021segformer} with patch\textunderscore size=3, and TransformerBlock as the efficient self-attention block from Segformer \cite{xie2021segformer} without dropout, with kqv\textunderscore bias, and with layer normalization.  Then the DeepLabV3 decoder features are fed to the module given in Fig. \ref{fig:E_low}, which outputs the low-resolution canonical features $\bm F$ as seen in the top half of the main paper's pipeline figure. Note that the output of this module is not technically $\bm F$, and instead $\bm F$ after being processed by additional convolutional layers.
\paragraph{Encoder for $\bm{F}_{\text{high}}$.}
 We then encode the image again (with its stacked pixel coordinates) with $\mathbf{E}_{\text{high}}$ with architectures given in \ref{fig:E_high}. Note that the input to the LT model's high-resolution encoder is the second layer output features of DeepLabV3, rather than the raw conditioning image. 
\paragraph{Final triplane encoder.}
Finally, $\bm F$ and $\bm F_\text{high}$ are concatenated and decoded to the triplane $\bm T$ with the architectures seen in Fig. \ref{fig:processor}, completing the final encoding stage seen in the main paper's pipeline figure.

\paragraph{Misc.}
For super-resolution, we used the same super-resolution network architecture as EG3D, but replaced the $w$ to be constant 1 as mentioned earlier. 
For volume rendering and decoding the triplane, we follow EG3D; specifically, we use 48 depth samples for coarse and fine passes for training. For discriminator, we use 2D dual-disciriminator from EG3D.

\paragraph{Training.}
In practice, we alternate between taking gradient steps with reference view supervision and multiview supervision. To do so, we begin by synthesizing synthetic input images for our encoder by sampling from the distribution for $\bm P_\text{ref}$ as detailed in the main paper. We render these cameras from triplanes from the frozen, pretrained EG3D. These are then fed to our encoder, whereby a triplane is then predicted. We can then render the same input cameras to take a gradient step for a loss computed only over the input views. We can additionally sample some cameras $\bm P_\text{mv}$, render ground truth from the EG3D triplanes, and render from the predicted triplanes for a multiview loss as well. In particular, at every gradient step, we always render 32 input cameras and 32 multiview cameras from the aforementioned distributions from the frozen EG3D. However, we do not always perform a gradient step for the input view loss.

In the first stage of the training, we compute losses for the reference set of cameras once every 10 triplane syntheses and perform gradient steps with respect to multiview supervision at every gradient step. We additionally do not incorporate any adversarial loss, nor category loss, and do not train the MLP decoder and super-resolution network parameters at all. We train for 30k iterations without these objectives in this first stage. In the second stage, we add the adversarial and category losses and backpropagate to all parameters in the pipeline, computing losses for the reference cameras every 2 gradient passes. In this stage, we remove the feature loss, and set the weight of the triplane loss to $0.01$. After 37.5k iterations, we begin to compute multiview supervision and reference view supervision at every EG3D triplane synthesis step and continued to reach 220k iters in total (including the first 37.5k). We use a learning rate of 1e-4 for the encoder parameters, except for the transformer parameters, which have a learning rate of 5e-5. We use the same settings as EG3D for the the discriminator. We train for about 10 days on 8 A100 GPUs or 8 A40 GPUs, for about 220k iterations in total. %

For training our model for cat faces, we used transfer learning following ~\cite{Karras2020ada,eg3d2022}. We initialized our cat face model with our human face model that is already trained, and ran training for additional 5 million images using the AFHQv2 checkpoint from EG3D.

\paragraph{Camera augmentation.}
EG3D assumes a fixed camera radius of 2.7, focal length of 18.83, zero camera roll, and a central principal point. For the FFHQ experiments, we sample the focal length from a normal distribution with standard deviation 1 centered at 18.83, the camera radius from a normal distribution centered at 2.7 with standard deviation 0.1, the principal point from a normal distribution with standard deviation 14 and centered at 256, and camera roll with a normal distribution of mean 0 and standard deviation 2 degrees.

For the AFHQ experiments, we sample the focal length from a normal distribution with standard deviation 1.5 centered at 18.83, the camera radius from a normal distribution centered at 2.7 with standard deviation 0.1, the principal point from a normal distribution with standard deviation 25 and centered at 256, and camera roll with a normal distribution of mean 0 and standard deviation 6 degrees.

\paragraph{Training data.}
We visualize the distribution of synthetic training data in two figures. Fig. \ref{fig:synthetic_pose} visualizes the limits of the input image poses in pitch and yaw for two subjects. Fig. \ref{fig:synthetic_focal} visualizes two sigmas of noise in the focal length and in the principal point used to augment camera information during our training.

\paragraph{Inference.}
To calculate the timings of our method, we wrap the forward calls of the encoder (not the rendering) in autocast, which we use for real-time applications. For renderings, we use 48 depth samples for real-time applications and 96 depth samples for the offline videos, following EG3D.

\section{Experiment details}
\label{sec:experiment_details}
\subsection{Baselines}
For all the baselines we used, we used official code from the authors with released pre-trained checkpoints. 

For HeadNeRF, we used the highest resolution model \textit{\texttt{model\_Reso64}} on the official website (\url{https://github.com/CrisHY1995/headnerf}), which produces the final output at 512 resolution using a feature map of resolution 64. 

For ROME, we use the pre-trained model from the official code release (\url{https://github.com/SamsungLabs/rome}), which produces the output at 256 resolution.

 For EG3D models, we used the FFHQ and AFHQv2 fine tuned models as described in Sec.~\ref{sec:implementation}, which are derived from the official EG3D models. The baseline "EG3D-PTI" combines the unconditional EG3D model with the lightweight generator tuning at test time using Pivotal Tuning Inversion (PTI) ~\cite{roich2021pivotal} for 3D GAN inversion from a single image. For the PTI inversion experiment, we follow the hyperparameter settings from the original PTI paper and the PTI experiment done in the EG3D paper, and optimize the latent code for 600 iterations, followed by fine tuning the generator weights for an additional 350 iterations. Unless noted otherwise, we used this setting for all our experiments. %

\paragraph{FFHQ}
For the comparisons on FFHQ between our model and other baselines, we postprocess our images with a rigid 2D warp. To accomplish this, we estimate 2D landmarks with an off-the-shelf facial landmark model \cite{bulat2017far} for both the ground truth and our predicted image. We then solve the Orthogonal Procrustes problem \cite{2020SciPy-NMeth} to find the optimal orthogonal matrix to rigidly align our image onto the target image approximately around the face region using the facial landmarks. Examples of this alignment can be seen in Fig. \ref{fig:misalignment}. We found that this alignment resulted in worse performance for EG3D-PTI and HeadNeRF, so we do not postprocess these methods' renderings. For comparison to ROME, we align our renderings and ROME's renderings to ROME's warped input (lower-resolution) image with the same process before computing the metrics. In any cases where the warp produced black pixels on the border (out of bounds), we set the ground truth, and the baselines pixels to black there as well, to ensure that we are comparing the exact same pixels between the methods. For ROME and HeadNeRF, we compare only on their valid pixels, using the provided masks from these methods.

To ensure fairness in the ID and Pose comparisons of our model against HeadNeRF and ROME, we postprocess our images to align with the output of each baseline. For HeadNeRF, we mask both our results and the ground truth results to the non-empty region using the provided HeadNeRF masks, and calculate the Pose and ID losses on the modified images. For ROME, we first downsample both our output and the ground truth images from $512^2$ to $256^2$ to align with the ROME output, then align the ROME output to the ground truth images using the same landmark detection and Procrustes alignment as described for PTI. Again, we mask both our output and ground truth to the non-empty region predicted by ROME, then calculate Pose and ID on the processed images.

\paragraph{H3DS.}
For the depth evaluations on the H3DS dataset \cite{ramon2021h3d}, we select a frontal image from all 23 subjects, then render the ground truth depth from the corresponding camera pose and the ground truth mesh. We normalize each depth to lie within [0,1]. We then feed the RGB images as input to all baselines, and compute each method's corresponding depth maps. For ours, ours (LT), EG3D-PTI, and ROME, we compute the scale- and translation-invariant L1 and RMSE errors only on the valid depth pixels from the ROME prediction. For HeadNeRF, we use only the valid depth pixels from its prediction. We found that the geometry of HeadNeRF can collapse to a plane in front of the predicted 3D face.

\section{Discussion}
\label{sec:discussion}
\paragraph{Ethical considerations.}
Since our method does not predict a latent space for portrait editing, it offers limited capabilities for portrait manipulations for malicious uses.   
However, it may be used to manipulate the viewpoint of a portrait. Potential solutions include detection of unseen image generators~\cite{stylegan3detector,Corvi2022diffusiondetection} and image watermarking~\cite{yu2022responsible, yu2021artificial}.

\paragraph{Adding real images to the training.}
Intuitively, incorporating real data into the training pipeline may be desirable in order to maximize the photorealism of rendered images and robustness in the most challenging settings. Future work may investigate the best way to use both synthetic and real data in conjunction with one another.

\paragraph{Extension to handling a video input.}
The framework of our model is such that we require only a single image at inference time. Our single-image method can be extended to handle a video input in a frame by frame fashion, but this may lead to flickering and temporal inconsistency when rendering videos due to the single-image nature of our model. In such cases where multiple images of a subject are available at inference, it is desirable to incorporate all such available information. Further work may investigate making the triplane autoregressive or recurrent, conditioned on the previous frames so that occlusions are handled in a consistent way, and there is greater temporal coherence.

{\small
\bibliographystyle{ieee_fullname}
\bibliography{egbib}
}